%% file: main.tex
\pdfoutput=1
\documentclass[10pt]{article} 
\usepackage[preprint]{tmlr}

\input{math_commands.tex}

\usepackage{hyperref}
\usepackage{url}

\usepackage{amssymb}            
\usepackage{mathtools}          
\usepackage{mathrsfs}           
\usepackage{graphicx}           
\usepackage{caption}
\usepackage{subcaption}         
\usepackage[space]{grffile}     
\usepackage{xcolor}             
\usepackage{tcolorbox}          
\usepackage{wrapfig}

\providecommand{\E}{\mathbb{E}}
\newtcolorbox[auto counter]{algbox}[2][]{title={Algorithm~\thetcbcounter: #2}, #1}

\title{Complexity-Regularized Proximal Policy Optimization}

\author{\name Luca Serfilippi \email luca.serfilippi@unibo.it \\
      \addr Department of Computer Science and Engineering\\
      Alma Mater Studiorum Università di Bologna
      \AND
      \name Giorgio Franceschelli \email giorgio.franceschelli@unibo.it \\
      \addr Department of Computer Science and Engineering\\
      Alma Mater Studiorum Università di Bologna
      \AND
      \name Antonio Corradi \email antonio.corradi@unibo.it \\
      \addr Department of Computer Science and Engineering\\
      Alma Mater Studiorum Università di Bologna
      \AND
      \name Mirco Musolesi \email m.musolesi@ucl.ac.uk \\
      \addr Centre for Artificial Intelligence, Department of Computer Science\\
      University College London\\
      and\\
      Department of Computer Science and Engineering\\
      Alma Mater Studiorum Università di Bologna
      }

\begin{document}

\maketitle

\begin{abstract}
Policy gradient methods usually rely on entropy regularization to prevent premature convergence. However, maximizing entropy indiscriminately pushes the policy towards a uniform distribution, often overriding the reward signal if not optimally tuned. We propose replacing the standard entropy term with a self-regulating complexity term, defined as the product of Shannon entropy and disequilibrium, which quantifies the distance from the uniform distribution. Unlike pure entropy, which favors maximal disorder, this complexity measure is zero for both fully deterministic and perfectly uniform distributions, i.e., it is strictly positive for systems that exhibit a meaningful interplay between order and randomness. These properties ensure the policy maintains beneficial stochasticity while reducing regularization pressure when the policy is highly uncertain, allowing learning to focus on reward optimization. We introduce Complexity-Regularized Proximal Policy Optimization (CR-PPO), a modification of PPO that leverages this dynamic. We empirically demonstrate that CR-PPO is significantly more robust to hyperparameter selection than entropy-regularized PPO, achieving consistent performance across orders of magnitude of regularization coefficients and remaining harmless when regularization is unnecessary, thereby reducing the need for expensive hyperparameter tuning.
\end{abstract}

\section{Introduction} \label{sec:introduction}

Reinforcement Learning (RL) has achieved significant results in numerous sequential decision-making problems, including board games \citep{silver2017mastering} and video games \citep{mnih2015humanlevel}, robotic control \citep{tang2025deep}, protein design \citep{angermueller2020modelbased}, and human alignment in language modeling \citep{ouyang2022training}. However, for certain real-world tasks, the state space might be incredibly vast, and the reward function too sparse, making agents prone to premature convergence to suboptimal deterministic policies. To counterbalance the convergent effect of reward maximization, research has mainly focused on inducing exploration through additional intrinsic rewards \citep{ladosz2022exploration} or regularizing policy learning by maximizing its entropy. Indeed, several policy gradient methods, such as Proximal Policy Optimization (PPO) \citep{schulman2017proximal}, often include \textit{entropy regularization}.
Entropy regularization encourages learning stochastic policies, helping with exploration, preventing deterministic behavior, and, in certain scenarios, facilitating the optimization process \citep{ahmed2019understanding}.
However, especially for PPO, its real utility is disputed, and finding the correct scaling factor for the entropy loss is not trivial \citep{andrychowicz2021what}; in addition, maximizing entropy pushes the policy towards a uniform random distribution regardless of the current nature of the policy and the actual need for exploration, which might result in a less efficient learning strategy \citep{zhang2025when}. The challenge is not merely to encourage exploration and prevent early convergence to local minima, but to regularize the policy \textit{robustly}. Intuitively, a regularizer should penalize convergence (determinism) without inhibiting learning by pushing continuously towards randomness. While defining \textit{complexity} is nuanced \citep{mitchell2009complexity}, 
physical definitions of complexity cover exactly this property. In particular, in statistical physics, a ``complex'' system is one that is neither perfectly ordered (crystal) nor perfectly random (ideal gas).

In this work, we propose a new regularization term based on complexity, defined as the product of Shannon entropy and disequilibrium, rather than the sole entropy. This measure, usually referred to as the L\'opez-Ruiz, Mancini, and Calbet (LMC) complexity measure \citep{lopezruiz1995statistical}, was developed to evaluate the complexity of physical systems at a given scale and is characterized by an interplay between the information stored by the system (its entropy) and the distance from equipartition (its disequilibrium). By acting as a regularizer, this term suppresses both extremes, thereby exerting pressure on agents to identify strategies that balance exploration and exploitation.
We then define \textbf{C}omplexity-\textbf{R}egularized \textbf{P}roximal \textbf{P}olicy \textbf{O}ptimization (\textbf{CR-PPO}), a new learning algorithm based on PPO that replaces its entropy bonus with a complexity bonus. By scaling the policy entropy with its disequilibrium, CR-PPO promotes divergence when the policy becomes too sharp and convergence when the policy is too flat. In addition, the proposed mechanism is inherently algorithm-agnostic and can be applied in the future to other policy optimization methods that rely on entropy-based regularization.
Experiments on several classic RL environments, as well as a novel version of CartPole with fine-grained complexity control named CARTerpillar,
demonstrate that CR-PPO is more robust to the choice of the loss scaling factor than entropy-regularized PPO and achieves competitive results across settings that require very different levels of exploration.

In summary, our main contributions are as follows:
\begin{itemize}
    \item We introduce a self-regulating complexity objective, based on the product of Shannon entropy and disequilibrium \citep{lopezruiz1995statistical}, to replace conventional entropy maximization in policy gradient methods. The proposed complexity measure vanishes for both deterministic and uniform distributions, thereby discouraging premature convergence to deterministic policies without indiscriminately driving the policy towards a uniform distribution.
    \item We introduce CR-PPO, a novel reformulation of PPO \citep{schulman2017proximal} that leverages a complexity maximization objective. We demonstrate empirically that CR-PPO is consistently more robust to hyperparameter choices than entropy-regularized PPO, while the proposed complexity regularization mechanism incurs no adverse effects when it is not needed.
    \item We present CARTerpillar, a variant of the classic CartPole environment with tunable difficulty via a single parameter, i.e., the number of carts interconnected with each other through dampers and springs, which influences both action and state spaces as well as the environmental dynamics, providing a systematic way to evaluate agent performance at increasing complexity.
\end{itemize}

\section{Related Work} \label{sec:related_work}

\subsection{Entropy Regularization in Reinforcement Learning} \label{sec:ent_regularization}

The concept of entropy has been extensively used in RL to address the exploration-exploitation trade-off \citep{sutton2018reinforcement}.
There are two primary strategies for optimizing entropy: the maximum entropy framework, in which policy entropy is incorporated as an intrinsic reward; and entropy regularization, in which policy entropy is optimized as a separate cost function.
The former has been studied in various RL domains, from inverse reinforcement learning \citep{ziebart2008maximum} to stochastic optimal control \citep{toussaint2009robot,rawlik2013stochastic} and especially off-policy algorithms \citep{haarnoja2017reinforcement,haarnoja2018soft}. The latter has been adopted in on-policy settings \citep{williams1991function,mnih2016asynchronous,schulman2017proximal} and in a combination of value-based and policy-based RL \citep{odonoghue2017combining}, and \citet{schulman2017equivalence} proved their equivalence under entropy regularization. While entropy regularization can make the policy optimization landscape smoother \citep{ahmed2019understanding}, entropy-regularized methods may fail to converge to a fixed point \citep{neu2017unified}. In addition, entropy is not always helpful \citep{liu2021regularization,zhang2025when}, and its coefficient plays a relevant role in its effectiveness \citep{andrychowicz2021what}.
Because of this, several modifications have been developed in the past. For example, \citet{zhao2019maximum} propose a reward-weighted entropy objective in the maximum entropy framework to solve multi-goal RL. In an off-policy setting, \citet{han2021maxmin} suggest training the Q-function to minimize entropy (visiting states with low entropy), while maintaining the policy entropy maximization term in the policy update. It is also possible to maximize the entropy of the weighted sum of the current policy action distribution and the sample action distribution from the replay buffer \citep{han2021diversity}, or to maximize the entropy of the state distribution induced by the current policy \citep{hazan2019provably}. In this work, we preserve the elegance and simplicity of (policy-based) entropy regularization, addressing its limitations by scaling it with the policy disequilibrium.

\subsection{Complexity and Reinforcement Learning} \label{sec:complexity_rl}

Recent research explores the idea that exposing models to structured complexity can lead to the development of more generalized and intelligent behaviors \citep{havrilla2024surveying}. 
\citet{zhang2025intelligence} investigate this phenomenon by pretraining large language models on synthetic data generated by elementary cellular automata, finding that models trained on data from systems at the ``edge of chaos'' \citep{langton1990computation}, poised between order and disorder, exhibit superior performance on downstream reasoning tasks.
Similarly, in the context of supervised learning, \citet{zhang2021edge} demonstrate that a network's generalization capability is maximized when its internal weight dynamics are poised at the critical boundary separating ordered and chaotic regimes. Their work shows how standard training algorithms naturally push the network towards this edge and how regularization can be precisely tuned to maintain the network in this optimal state.
Complexity plays a key role in reinforcement learning as well. To enhance the robustness and stability of the learned policy, \citet{young2024enhancing} use the Maximal Lyapunov Exponent (MLE; \citealp{lyapunov1992general}) from chaos theory to demonstrate that policies can be highly sensitive to initial conditions, where small perturbations in the state lead to vastly different long-term trajectories. To mitigate this instability, they propose a novel regularization term that directly penalizes the MLE, encouraging the agent to learn more stable and predictable dynamics.
Conversely, our work builds on the idea that complexity-based regularization should not aim for predictable dynamics, but for more stable and less random exploration.

\section{Background} \label{sec:background}

\subsection{Reinforcement Learning and Proximal Policy Optimization} \label{sec:ppo}

We focus on Markov Decision Processes expressed by the tuple $\left(\mathcal{S}, \mathcal{A}, P, r, \gamma\right)$, where $\mathcal{S}$ is the state space, $\mathcal{A}$ is the discrete action space, $P(s' | s, a)$ is the transition probability between states, $r(s, a)$ is the reward associated with that transition, and $\gamma$ is the discount factor. The RL agent uses a $\bm{\theta}$-parameterized policy $\pi_{\bm{\theta}}(a|s)$, i.e., a mapping from states to distributions over actions, to maximize the discounted return $\E_{\pi_{\bm{\theta}}}[\sum_{t=0}^T \gamma^t r(s_t, a_t)]$.
Many policy gradient methods learn $\pi_{\bm{\theta}}$ via stochastic gradient ascent over the estimator of the gradient provided by the policy gradient theorem \citep{sutton1999policy}: $\nabla_{\bm{\theta}}J(\bm{\theta}) \propto \E_{\pi_{\bm{\theta}}}[\nabla_{\bm{\theta}} \log \pi_{\bm{\theta}}(a_t | s_t) \hat{A}_t]$, where $\hat{A}_t$ is an estimator of the advantage function at timestep $t$, which quantifies, given $s_t$, how much better $a_t$ is with respect to other actions on average.
Several policy gradient algorithms have been derived from the policy gradient theorem. However, directly updating the policy by a step of gradient ascent through a sample-based estimate leads to poor data efficiency and instability. Proximal Policy Optimization (PPO) addresses these issues by constraining the update size with a clipped surrogate objective that provides a stable and reliable update and allows for reusing samples over multiple updates \citep{schulman2017proximal}:
\begin{equation} \label{eq:ppo_clip}
    L^{CLIP}_t\!\left(\bm{\theta}\right) = \min\!\left(\frac{\pi_{\bm{\theta}}\!\left(a_t | s_t\right)}{\pi_{\bm{\theta}_{old}}\!\left(a_t | s_t\right)} \hat{A}_t, \mathrm{clip}\!\left(\frac{\pi_{\bm{\theta}}\!\left(a_t | s_t\right)}{\pi_{\bm{\theta}_{old}}\!\left(a_t | s_t\right)}, 1 - \epsilon, 1 + \epsilon\right) \hat{A}_t\right),
\end{equation}
\noindent where $\pi_{\bm{\theta}_{old}}$ is the policy before the update from which the action has been sampled, and $\epsilon$ is the clipping factor. The advantage $\hat{A}_t$ is computed as the generalized advantage estimation \citep{schulman2016highdimensional} and makes use of a state-value function approximator $V_{\bm{\theta}}$ trained to minimize $L^{VF}_t(\bm{\theta}) = \hat{\mathop{\mathbb{E}}}_t \!\left[( V_{\bm{\theta}}(s_t) - V^{targ}_t)^2 \right]$.
Finally, PPO also includes an entropy bonus $H[\pi_{\bm{\theta}}]$ to ensure sufficient exploration; thus, the overall objective becomes:
\begin{equation} \label{eq:entropy_obj}
    L(\bm{\theta})=\hat{\E}_{t}\left[L_{t}^{CLIP}(\bm{\theta}) - c_{vf} L_{t}^{VF}(\bm{\theta}) + c_{reg} H[\pi_{\bm{\theta}}](s_t)\right],
\end{equation}
\noindent with $c_{vf}, c_{reg}$ coefficients for the value-function loss and the entropy term, respectively.

\subsection{LMC Complexity} \label{sec:lmc}

L\'opez-Ruiz, Mancini, and Calbet (LMC) complexity is a measure based on a probabilistic description of physical systems \citep{lopezruiz1995statistical}. Unlike other complexity measures that analyze streams of data (e.g., \citealp[]{grassberger1986toward, lempel1976complexity}), it is based on the statistical description of systems at a given scale, and on the idea that a system is said to be \textit{complex} when it does not conform to patterns regarded as \textit{simple}. Two physical systems have simple models and, thus, ideal zero complexity: the perfect crystal, which is completely ordered and can be described with minimal information; and the isolated ideal gas, which is completely disordered and has maximal information. In contrast, all other particle systems exhibit non-zero complexity. Similarly, in RL, two policies correspond to simple distributions with ideal zero complexity: a fully random policy and a deterministic policy. All other policies exhibit non-zero complexity, with increasing complexity when all actions are feasible but certain actions have a greater probability, that is, when the policy maintains a sharp yet stochastic nature. Assuming a system with $N$ accessible states $\{x_1, x_2, \ldots, x_N\}$, each with probability $p_i$ with $\sum_{i=1}^N p_i = 1$, the LMC complexity is defined as:
\begin{equation} \label{eq:lmc_complexity}
    C = H \cdot D = \underbrace{\left(- \sum_{i=1}^N p_i \log p_i\right)}_\textrm{Entropy} \cdot \underbrace{\left(\sum_{i=1}^N \left(p_i - \frac{1}{N}\right)^2\right)}_\textrm{Disequilibrium}.
\end{equation}
In essence, LMC complexity is characterized by an interplay between the information stored by the system and the distance from equipartition. Thus, it is equal to zero if the entropy is zero (as for the perfect crystal or, in the case of RL, when an action has a probability of $1$) or if the disequilibrium is zero (as for the isolated ideal gas or, in the case of RL, when all actions are equiprobable); and it is high when the system is both stochastic and sharp.

\section{Method} \label{sec:c3po}

\begin{wrapfigure}{rt}{0.5\textwidth}
\vspace{-12pt}
    \centering
    \includegraphics[width=0.85\linewidth]{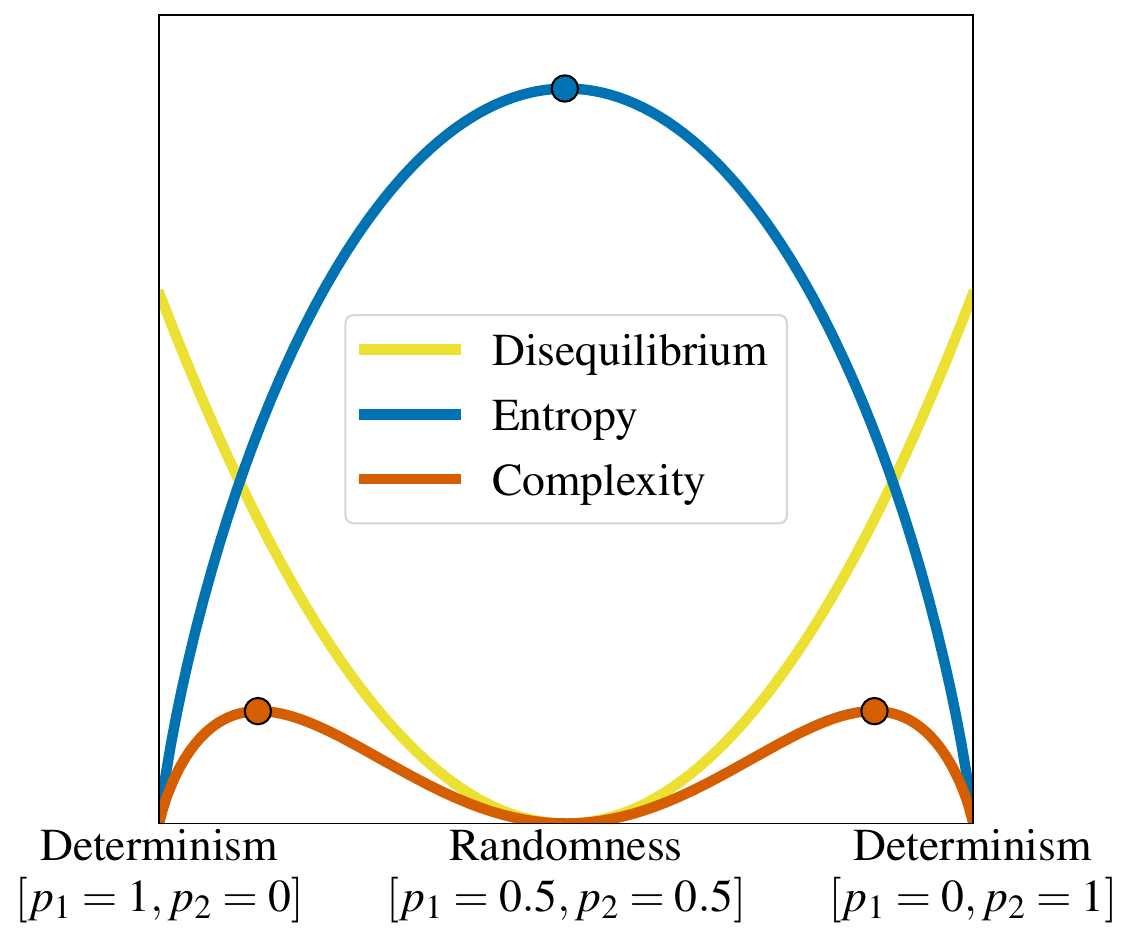}
    \caption{Disequilibrium, entropy, and complexity in a two-dimensional space. The bullet points indicate their maxima.}
    \label{fig:2actioncomplexity}
\vspace{-12pt}
\end{wrapfigure}

Incorporating entropy maximization into policy gradient methods, such as PPO, has been shown to enhance exploration and prevent the agent from converging to deterministic strategies. However, maximizing entropy unconditionally drives the policy towards a uniform distribution. If the entropy scaling factor is too high, the entropy term can dominate the policy loss, preventing the agent from finding an optimal solution. Furthermore, in environments that do not require extensive exploration, excessive entropy can unnecessarily slow convergence \citep{zhang2025when}.

To avoid aiming at randomness too much, but still preventing determinism, we propose replacing the entropy loss with an LMC complexity loss. As introduced in Section \ref{sec:lmc}, LMC complexity aims to avoid \textit{simplicity}; the system must have many possible states (high entropy), and, at the same time, it must not be completely random (high disequilibrium). While entropy is maximum if all events are equiprobable and minimum if only one of them is possible, complexity is minimum if either all events are equiprobable or only one of them is possible, and has multiple maxima in between. Figure \ref{fig:2actioncomplexity} illustrates its behavior for a system with 2 possible events: whenever the system is too sharp, maximizing complexity pushes it towards randomness; and whenever the system is too random, maximizing complexity pushes it towards sharpness.

More formally, given a discrete action space $\mathcal{A}$ and a policy $\pi_{\bm{\theta}}(a|s)$, the complexity of the policy can be defined as follows:
\begin{equation} \label{eq:complexity_policy}
    C[\pi_{\bm{\theta}}](s) = H[\pi_{\bm{\theta}}](s) \cdot D[\pi_{\bm{\theta}}](s) = \underbrace{ - \sum_{a \in \mathcal{A}}{\pi_{\bm{\theta}}(a|s)\log \pi_{\bm{\theta}}(a|s)} }_\textrm{Entropy} \cdot \underbrace{ \sum_{a \in \mathcal{A}}{\left(\pi_{\bm{\theta}}(a|s) - \frac{1}{|\mathcal{A}|}\right)^2} }_\textrm{Disequilibrium}.
\end{equation}
This quantity can then be optimized for by any policy gradient method in order to foster complexity rather than the sole entropy. In particular, we define here a new learning algorithm, namely Complexity-Regularized Proximal Policy Optimization (CR-PPO), which leverages the PPO learning scheme while multiplying the entropy with the policy disequilibrium, i.e., replacing its entropy term $H[\pi_{\bm{\theta}}]$ with Equation \ref{eq:complexity_policy}. In this way, CR-PPO moves from trying to randomize the policy to trying to maintain stochasticity without falling into the randomness pit. Overall, the objective function of CR-PPO is as follows: 
\begin{equation} \label{eq:complexity_obj}
    L(\bm{\theta})=\hat{\E}_{t}\left[L_{t}^{CLIP}(\bm{\theta}) - c_{vf} L_{t}^{VF}(\bm{\theta}) + c_{reg} C\left[\pi_{\bm{\theta}}\right]\!(s_{t})\right].
\end{equation}
An analysis of its gradient and the benefits it brings is reported in Appendix \ref{gradient_analysis}.

From a practical standpoint, we can extend current PPO implementations to CR-PPO and promote complexity rather than randomness by just computing the value of the disequilibrium and multiplying it by the entropy. This incurs negligible computational overhead and requires no architectural changes. Similarly, this complexity-based mechanism can be applied to any other policy approximation algorithm that relies on entropy-based regularization. It is important to note that the current disequilibrium formulation applies only to discrete action spaces. However, it is possible to foresee different formulations that can be extended to continuous domains, e.g., by considering the variance magnitude and/or computing the integration rather than the summation.

\begin{algbox}[label={alg:crppo}, colback=gray!10, colframe=black!80, left=2pt, top=2pt, bottom=2pt, boxsep=2pt, boxrule=1.0pt]{Complexity-Regularized Proximal Policy Optimization}
Input: Initialized network parameters $\bm{\theta}$, step size $\mu$\\
Loop for iteration = $1, 2, \ldots$ :\\
\hspace*{1.5em}Collect a set of trajectories $\mathcal{D} = \{s_i, a_i, \hat{A}_i, V_i^{targ}, \pi_{\bm{\theta}_{old}}(a_i|s_i)\}_{i=1}^D$ following $\pi_{\bm{\theta}}(\cdot | \cdot)$\\
\vspace{1mm}\hspace*{1.5em}Loop for each policy epoch = $1 \ldots N$:\\
\hspace*{3em}Split the training batch $\mathcal{D}$ into $M = \frac{|\mathcal{D}|}{B}$ random minibatches\\
\vspace{1mm}\hspace*{3em}Loop for each minibatch $k = 1 \ldots M$:\\
\vspace{1mm}\hspace*{4.5em}$\hat{A}_t = \frac{\hat{A}_t - \mathrm{mean}(\{\hat{A}_i\}_{i=1}^B)}{\mathrm{std}(\{\hat{A}_i\}_{i=1}^B)}$ for $t = 1 \ldots B$\\
\vspace{1mm}\hspace*{4.5em}$C[\pi_{\bm{\theta}}](s_t) = - \sum_{a \in \mathcal{A}} \pi_{\bm{\theta}}(a | s_t) \log \pi_{\bm{\theta}}(a | s_t) \cdot \sum_{a \in \mathcal{A}} (\pi_{\bm{\theta}}(a | s_t) - \frac{1}{|\mathcal{A}|})^2$ for $t = 1 \ldots B$\\
\vspace{1mm}\hspace*{4.5em}$L_t^{CLIP}\!\left(\bm{\theta}\right) = \min\!\left(\frac{\pi_{\bm{\theta}}\!\left(a_t | s_t\right)}{\pi_{\bm{\theta}_{old}}\!\left(a_t | s_t\right)} \hat{A}_t, \mathrm{clip}\!\left(\frac{\pi_{\bm{\theta}}\!\left(a_t | s_t\right)}{\pi_{\bm{\theta}_{old}}\!\left(a_t | s_t\right)}, 1 - \epsilon, 1 + \epsilon\right) \hat{A}_t\right)$ for $t = 1 \ldots B$\\
\vspace{1mm}\hspace*{4.5em}$L_t^{VF}(\bm{\theta}) = ( V_{\bm{\theta}}(s_t) - V^{targ}_t)^2$ for $t = 1 \ldots B$\\
\vspace{1mm}\hspace*{4.5em}$L(\bm{\theta}) = \hat{\mathop{\mathbb{E}}}_t \!\left[ L_t^{CLIP}(\bm{\theta}) - c_{vf} L_t^{VF}(\bm{\theta}) + c_{reg} C[\pi_{\bm{\theta}}](s_t) \right]$\\
\vspace{1mm}\hspace*{4.5em}$\bm{\theta} \leftarrow \bm{\theta} + \mu \nabla_{\bm{\theta}} L(\bm{\theta})$
\end{algbox}

\section{Experiments} \label{sec:experiments}

This work focuses on regularization as opposed to developing separate exploration mechanisms. 
While several methods generate auxiliary rewards to encourage exploration \citep{ladosz2022exploration}, entropy (and, by extension, complexity) regularization modifies the policy objective directly to shape the stochasticity of the action selection. As such, CR-PPO is orthogonal to auxiliary reward methods and can be combined with them to improve exploration while preserving its regularization benefits.
To rigorously isolate the contribution of the regularization objective and assess its hyperparameter robustness, our experiments consider PPO, comparing both entropy- and non-regularized configurations as baselines.\footnote{The code is available at: \url{https://github.com/Lucaserf/CR-PPO}.}

\subsection{Experimental Setup}

We evaluate CR-PPO on a diverse suite of environments with discrete, variable-length action spaces that have been shown to require different degrees of entropy regularization: CartPole (2-dimensional action space) and CarRacing (5), along with the Atari games Asteroids (14) and Riverraid (18), all taken from \texttt{gymnasium} \citep{towers2025gymnasium}; CoinRun (15) from \texttt{procgen} \citep{cobbe2020leveraging}; and Crafter (17) \citep{hafner2022benchmarking}.
This set of environments enables testing whether CR-PPO correctly modulates regularization across the full spectrum of requirements, from minimal to crucial.
We compare CR-PPO against two key baselines: i) PPO with an entropy bonus (from hereon, PPOwEnt), and ii) PPO without entropy (PPOwoEnt) to establish the floor performance and demonstrate the need for regularization in harder environments. For both CR-PPO and PPOwEnt, we vary the regularization coefficient $c_{reg}$ to evaluate their robustness, and we test the values $[1\text{e-}1, 5\text{e-}2, 1\text{e-}2, 5\text{e-}3, 1\text{e-}3]$ across all benchmark environments.
We repeat all the experiments across 5 different seeds. The architectures and hyperparameters used are taken from \texttt{RL-Baselines3-Zoo} \citep{raffin2021rl} and reported in Appendix \ref{sec:implementation_details}.

\subsection{Experimental Results}

\begin{figure}[t]
    \centering
    \includegraphics[width=1\linewidth]{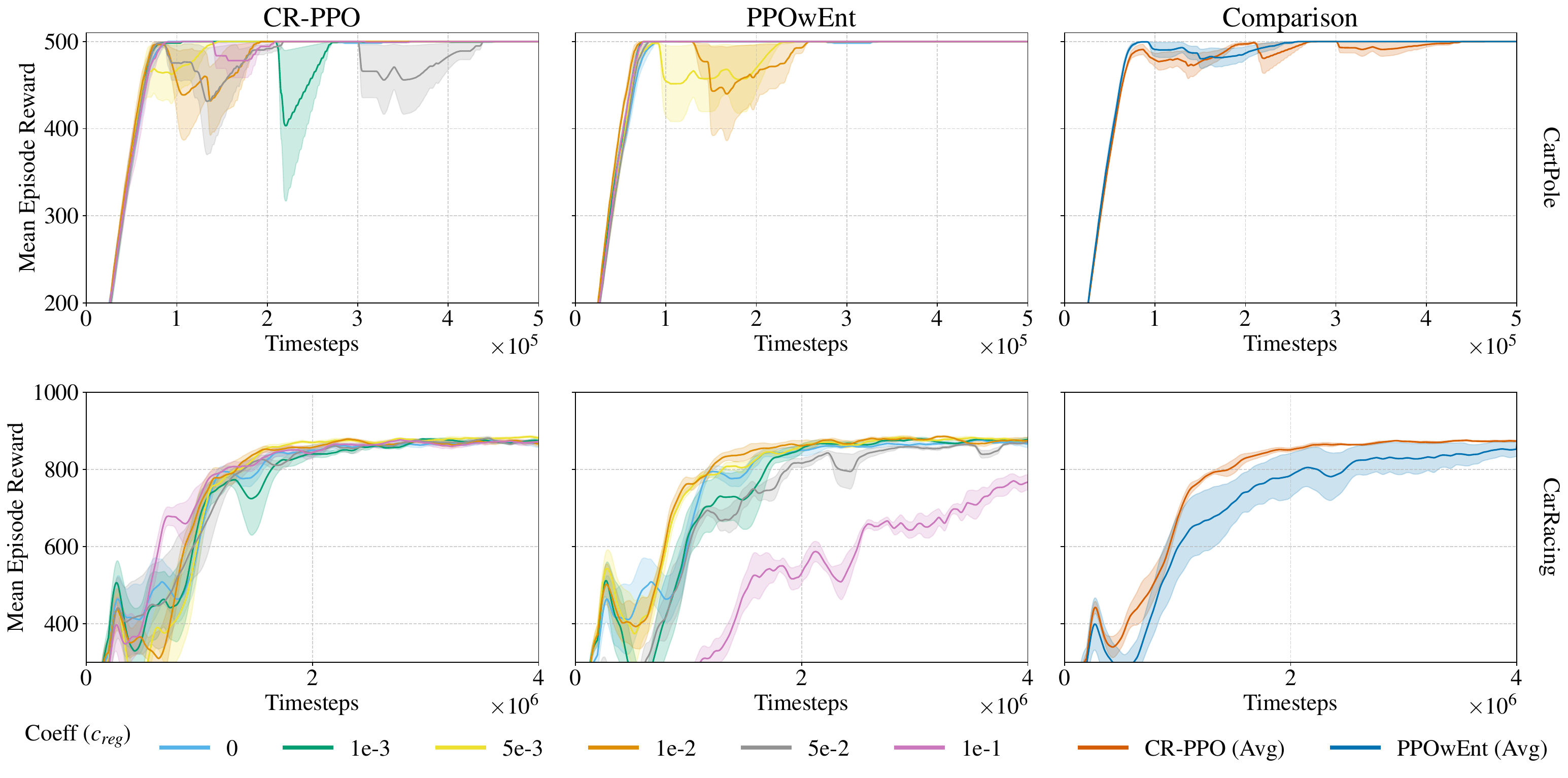}
    \caption{CartPole (top) and CarRacing (bottom) mean return for different $c_{reg}$ values of CR-PPO (left) and PPOwEnt (center), and their aggregated average (right). The mean and standard error are shown across 5 seeds.}
    \label{fig:cartpole_carracing}
\end{figure}

\begin{figure}[ht]
    \centering
    \includegraphics[width=1\linewidth]{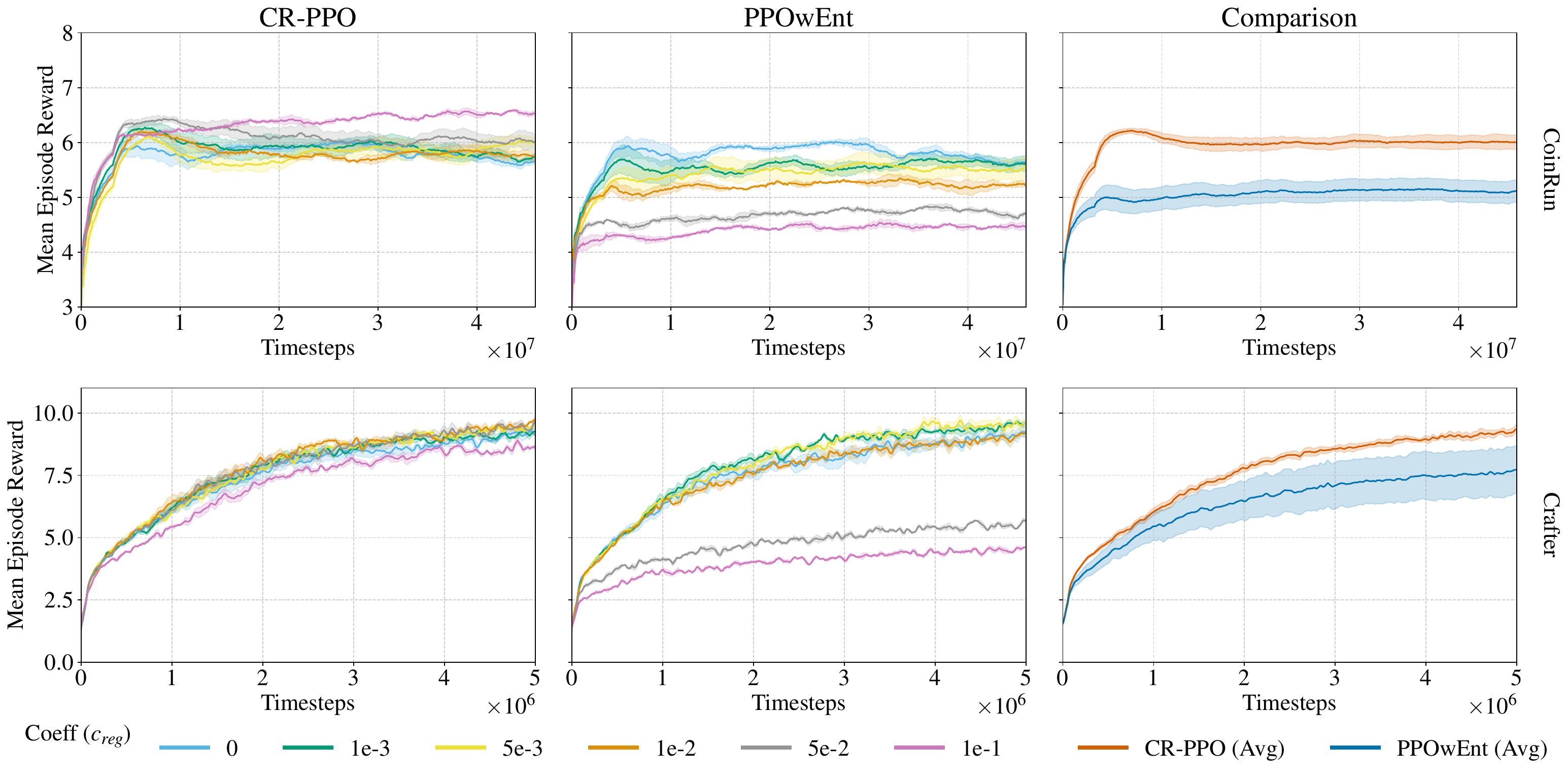}
    \caption{CoinRun (top) and Crafter (bottom) mean return for different $c_{reg}$ values of CR-PPO (left) and PPOwEnt (center), and their aggregated average (right). The mean and standard error are shown across 5 seeds.}
    \label{fig:coinrun_crafter}
\end{figure}

\begin{figure}[ht]
    \centering
    \includegraphics[width=1\linewidth]{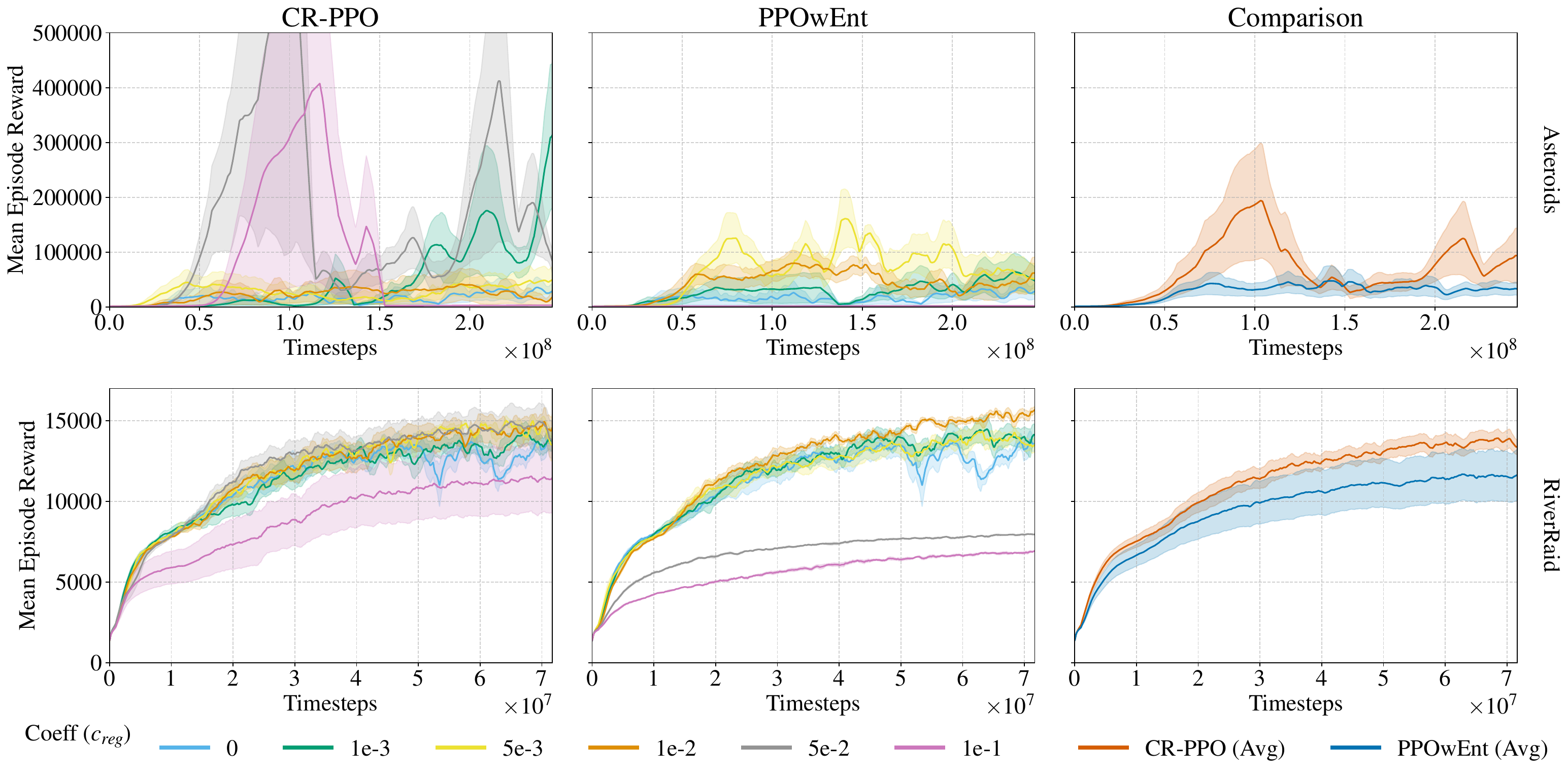}
    \caption{Asteroids (top) and Riverraid (bottom) mean return for different $c_{reg}$ values of CR-PPO (left) and PPOwEnt (center), and their aggregated average (right). The mean and standard error are shown across 5 seeds.}
    \label{fig:asteroids_riverraid}
\end{figure}

The results of the experiments are reported in 
Figures \ref{fig:cartpole_carracing} to \ref{fig:asteroids_riverraid}.
Each figure shows the learning curves for CR-PPO (left) and PPOwEnt (center) under different coefficient values (including without entropy), and also reports a comparison of their aggregated performance across all non-zero coefficients (right).
Our experiments reveal three distinct patterns regarding the role of the regularization term.

\textbf{Environments where entropy regularization has minimal impact.} In simpler tasks like CartPole and CarRacing (Figure \ref{fig:cartpole_carracing}), regularization does not impact the final results. Here, both CR-PPO and PPOwEnt perform on par with PPOwoEnt, demonstrating that the complexity bonus does not hinder performance when it is not needed. However, very high entropy coefficients can slightly slow down PPOwEnt learning, a sensitivity not observed with CR-PPO.

\textbf{Environments where entropy regularization is detrimental.} In environments like CoinRun (Figure \ref{fig:coinrun_crafter}, top), aggressive regularization is counterproductive. This might be because only a small percentage of available actions is actually helpful; blind regularization forces the agent to select ineffective actions, preventing (fast) convergence. Indeed, increasing the entropy coefficient severely degrades PPO performance. In contrast, CR-PPO remains robust across all coefficient values, consistently matching or improving upon the baseline performance by avoiding an overly random policy. A similar pattern is visible in Crafter (Figure \ref{fig:coinrun_crafter}, bottom), where high entropy coefficients stall learning, while CR-PPO maintains stability over a wider range of values.

\textbf{Environments where entropy regularization is beneficial.} For more complex tasks like Asteroids and Riverraid (Figure \ref{fig:asteroids_riverraid}), introducing regularization is key to achieving high scores.
Here, PPOwoEnt is sub-optimal or even incapable of maximizing the reward. While an optimally-tuned entropy bonus can improve performance over the baseline, PPOwEnt is highly sensitive to the coefficient value $c_{reg}$: a too-high value deteriorates performance, while a too-low value is ineffective. CR-PPO achieves comparable or superior results to a well-tuned entropy bonus but does so across a much broader range of coefficients.

\subsection{CARTerpillar}\label{subsec:carterpillar}

\begin{figure}[ht]
    \centering
    \includegraphics[width=1\linewidth]{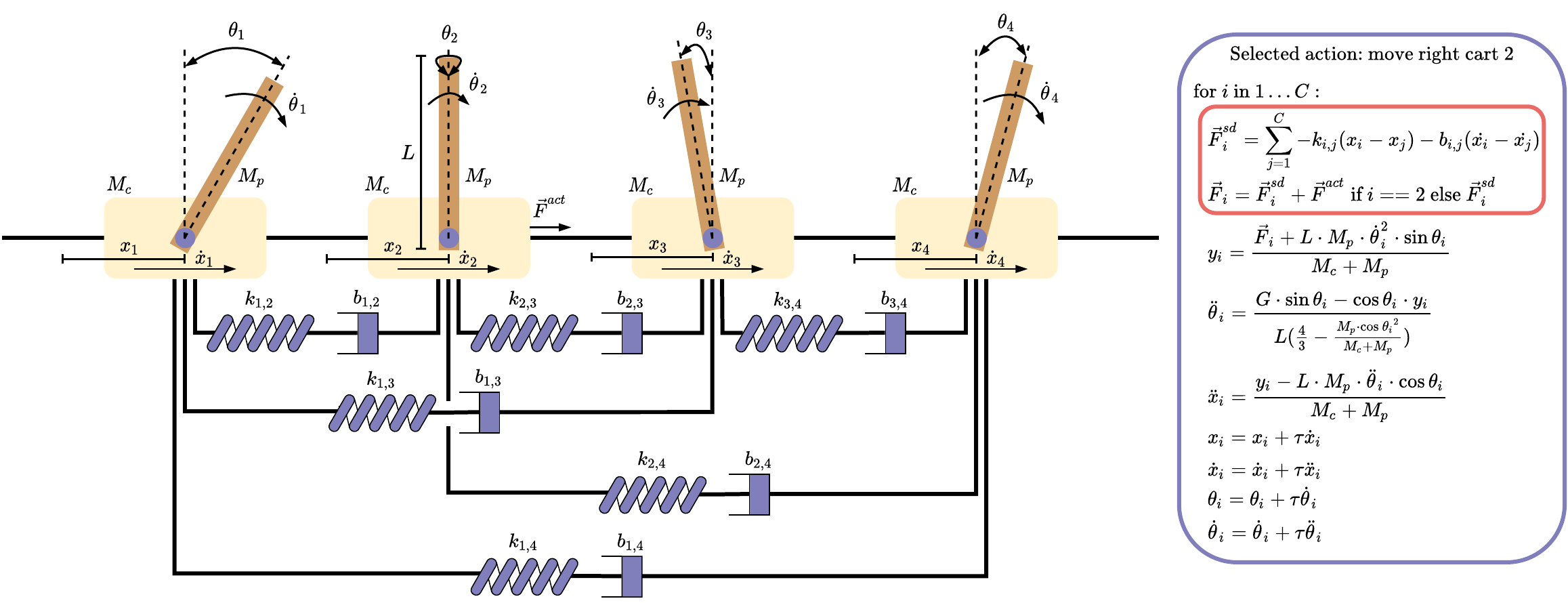}
    \caption{Schematic representation of the CARTerpillar environment with 4 carts. On the left, a flattened rendering with all the relevant physical symbols. On the right, the full dynamics of the environment, where the effect of newly introduced dampers and springs is highlighted in red.}
    \label{fig:carterpillar_summary}
\end{figure}

To systematically evaluate how regularization affects performance as task complexity increases, we designed a new environment with tunable difficulty. Standard benchmarks are often fixed, making it difficult to observe the varying necessity of regularization strategies. Our proposed environment, which we name CARTerpillar\footnote{A \texttt{gym}-based implementation can be found at: \url{https://github.com/Lucaserf/CARTerpillar}.}, is an extension of the classic CartPole environment to multiple carts, where the number of carts $K$ to be balanced is a parameter that controls the difficulty of the environment. The $K$ carts form a fully-connected system, with all possible pairs of carts $i, j$ linked by a damper (with a constant factor of $b_{i,j} = b$) and a spring (with a constant factor of $k_{i,j} = k$). 
Adding an $i$-th cart increases the number of connections by $i-1$, making the dynamics more challenging, and expands the discrete action space by two actions (push left/right for the new cart). At each timestep, the agent can only move one cart left or right; however, each cart is also subject to a force that is due to all the dampers and springs connecting them. Figure \ref{fig:carterpillar_summary} reports a simplified rendering of CARTerpillar, together with its full physical dynamics.

\begin{figure}[ht]
    \centering
    \includegraphics[width=1.\linewidth]{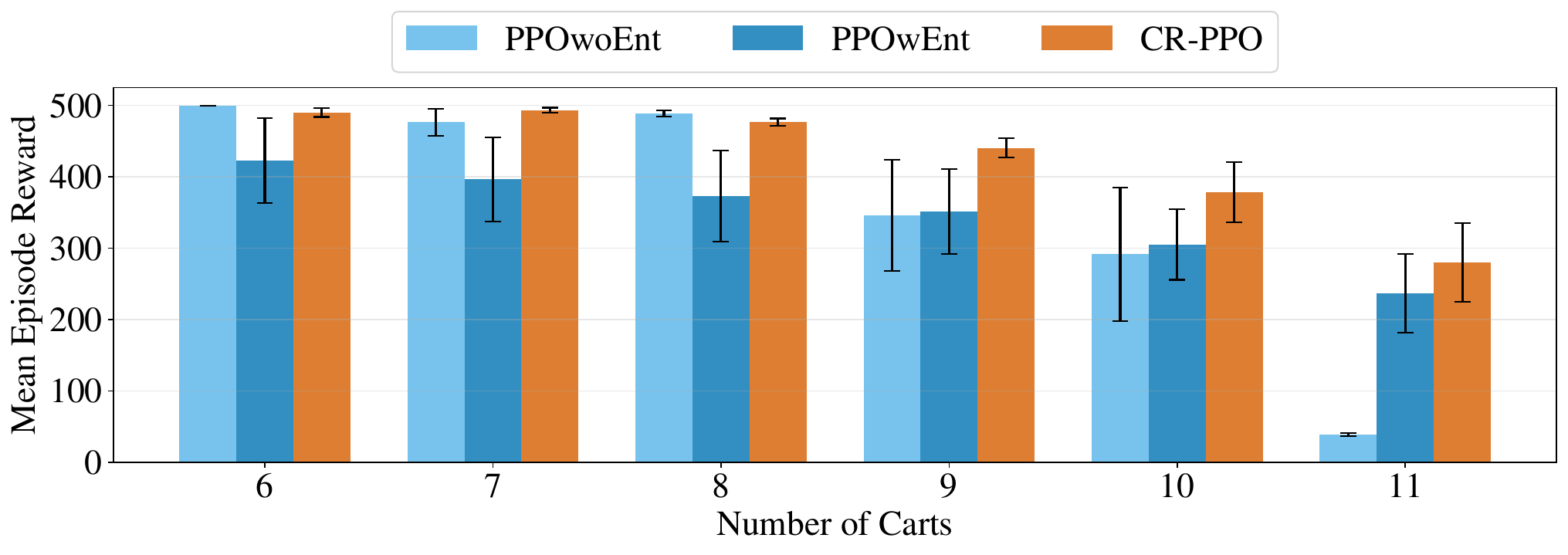}
    \caption{CARTerpillar aggregated results. Each bar reports the aggregated mean episode reward and the standard error across five seeds and all tested $c_{reg}$ coefficients.}
    \label{fig:carterpillar_results}
\end{figure}

We compare CR-PPO against PPO with and without entropy regularization (with $[1\text{e-}1, 3\text{e-}2, 1\text{e-}2, 3\text{e-}3, 1\text{e-}3]$ $c_{reg}$ values) over CARTerpillar across different $K$ values. As summarized in Figure \ref{fig:carterpillar_results}, all methods converge to the optimum for $K < 9$, whereas for $K \in [9, 10, 11]$, the aggregated results clearly demonstrate the relationship between task difficulty and the utility of regularization. As the number of carts increases, the performance of the baseline PPOwoEnt drops significantly, highlighting the growing need for regularization. While entropy regularization is beneficial for harder configurations (e.g., 10 and 11 carts), our complexity-based approach proves to be more robust to different $c_{reg}$ values (see Appendix \ref{sec:additional_results} for a detailed analysis). To sum up, CR-PPO does not hinder performance in simpler settings and consistently outperforms a non-optimally tuned PPOwEnt in the more challenging, high-dimensional scenarios.

\section{Discussion} \label{sec:discussion}

Overall, our experimental results demonstrate that CR-PPO offers a more stable and reliable alternative to entropy-regularized PPO. It effectively adapts the level of regularization pressure, proving beneficial in complex environments while remaining harmless in simpler ones, thereby reducing the need for meticulous hyperparameter tuning. Specifically, CR-PPO encourages sufficient exploration when necessary, maintaining a stochastic (but not random) policy regardless of the scaling coefficient, while entropy requires tuning to reach the same performance. When regularization is unnecessary, as in simpler environments, its inclusion does not negatively affect performance, and CR-PPO performs on par with the non-regularized PPO across all tested coefficients for the complexity term. This holds even in cases where regularization is harmful, such as when a policy must become occasionally deterministic to make precise decisions. Unlike entropy, complexity has a more nuanced landscape, where solutions with near-zero probabilities in certain dimensions can still lie near an optimum. In other words, a policy with some near-zero probability actions may have low entropy, yet still retain reasonable complexity. This distinction ensures that complexity-based regularization remains effective even in scenarios where entropy regularization degrades performance.

In summary, complexity maximization emerges as a strong regularization method, especially in settings where the appropriate level of regularization is unknown. Its robustness to the choice of $c_{reg}$ can drastically reduce the need for parameter tuning, with substantial savings in energy consumption and computational cost. However, the $c_{reg}$ coefficient still influences the final result, though less significantly than for entropy. Additionally, our evaluation is limited to classic RL scenarios with a relatively small action space. Whether the observed benefits scale up with the number of actions is an open question. Finally, due to the mathematical formulation of disequilibrium, CR-PPO currently applies only to environments with discrete action spaces. Extending our approach to continuous actions represents a compelling direction for future work.

\section{Conclusion} \label{sec:conclusion}

We introduced Complexity-Regularized Proximal Policy Optimization (CR-PPO), a variant of PPO with a regularization strategy that replaces the monotonic push of entropy maximization with a distribution-aware complexity objective. 
Our theoretical analysis suggests that CR-PPO acts as an auto-tuning regularizer: it provides high exploration incentives when the policy becomes too sharp but discourages randomness as the policy becomes too flat.
We have conducted extensive experiments on classical reinforcement learning environments and CARTerpillar, a modified CartPole that allows for fine-grained control of its difficulty, finding that, especially for harder tasks, complexity improves upon entropy regularization by being more resilient to its scaling factor and providing better results in more complex environments.
Our research agenda includes extending the scope of complexity regularization to continuous action spaces and investigating its interplay with off-policy algorithms, as well as combining it with curiosity-driven strategies to enhance further exploration. Additionally, we will explore the application of complexity regularization in more practical scenarios, such as language modeling and decision-making, to further stabilize reinforcement learning algorithms in real-world settings.

\subsubsection*{Broader Impact Statement}
In this work, we propose an algorithmic contribution to reinforcement learning policy regularization. We do not anticipate immediate direct negative ethical consequences from this theoretical and algorithmic framework. Nevertheless, as reinforcement learning techniques are deployed in autonomous decision-making systems, robotics, and critical domains, ensuring reliable, robust, and safe exploration remains essential. The authors encourage practitioners to carefully assess application-specific safety constraints when applying policy optimization algorithms.

\bibliography{biblio}
\bibliographystyle{tmlr}

\appendix

\section{Gradient Analysis} \label{gradient_analysis}

To explain why the complexity bonus is more robust than the entropy bonus, we now examine their respective gradients. We will show that the complexity optimization landscape contains multiple maxima. Furthermore, unlike the entropy bonus, it actively lowers the probability of both deterministic and purely random behaviors.

Given the probability distribution $\mathbf{p} = \left(p_1, p_2, \ldots, p_{|\mathcal{A}|}\right)$ with decision variables $p_i = \pi_{\bm{\theta}}(a_i | s)$ outputted by the $\bm{\theta}$-parameterized neural network, the gradient of the entropy bonus can be written as $\nabla H[\mathbf{p}] = - \sum_{i=1}^{|\mathcal{A}|} \left(\log p_i + 1\right) \nabla p_i$ subject to the probability constraint $\sum_{i=1}^{|\mathcal{A}|} p_i = 1$. Under this constraint, the maximum lies in $p_i = \frac{1}{|\mathcal{A}|}, \forall \, i \in \left[1, |\mathcal{A}|\right]$ (the proof is reported in Appendix \ref{sec:proofs}), and optimization drives the policy towards a uniform distribution: if $p_i > \frac{1}{|\mathcal{A}|}$, the sign is negative, and the $i$-th probability is decreased; instead, if $p_i < \frac{1}{|\mathcal{A}|}$, the sign becomes positive, and the probability is increased.
Conversely, the gradient of the disequilibrium term is $\nabla D[\mathbf{p}] = \sum_{i=1}^{|\mathcal{A}|} 2\left(p_i - \frac{1}{|\mathcal{A}|}\right) \nabla p_i$, whose sign is positive when the action probability is greater than $\frac{1}{|\mathcal{A}|}$ and negative when it is lower, thus pushing the distribution even further away from equilibrium. The gradient of both terms can then be extended to $\bm{\theta}$ parameters by the chain rule.
Finally, the complexity bonus $C[\mathbf{p}] = H[\mathbf{p}] \cdot D[\mathbf{p}]$ has the following gradient given by the product rule:
\begin{equation}
    \nabla C[\mathbf{p}] = \sum_{i=1}^{|\mathcal{A}|} \left[2H[\mathbf{p}]\left(p_i - \frac{1}{|\mathcal{A}|}\right) - D[\mathbf{p}] \left(\log p_i + 1\right) \right] \nabla p_i,
\end{equation}
\noindent and has $|\mathcal{A}|$ maxima (see Appendix \ref{sec:proofs}). The term in brackets dictates the update and behaves less straightforwardly, depending on the current policy. When the policy is near-deterministic, i.e., $H \approx 0$, the second term dominates the gradient, making the policy flatter. Conversely, when the policy is near-uniform, i.e., $D \approx 0$, the first term dominates the gradient, encouraging the policy to become sharper. Finally, when both entropy and disequilibrium are large ($H, D \gg 0$), both terms influence the gradient sign, creating a dynamic equilibrium that keeps the policy stochastic without collapsing into a simple, deterministic strategy or dissolving into pure randomness.
Overall, this self-regulating mechanism, which encourages exploration when the policy is nearly deterministic and pushes toward sharpness when it becomes too flat, offers a theoretical explanation for the stability and robustness of CR-PPO observed in our experiments regardless of the actual scaling factor.
This highlights the interplay between exploration and exploitation that enables consistent learning and prevents collapse into trivial or random policies.

\section{Proofs} \label{sec:proofs}

Entropy is defined as $H[\mathbf{p}] = - \sum_{i=1}^{|\mathcal{A}|} p_i \log p_i$ and subject to the probability constraint $\sum_{i=1}^{|\mathcal{A}|} p_i = 1$. To compute its gradient subject to the constraint, we can make use of the method of Lagrange multipliers, where the gradient of $f(x)$ subject to $g(x) = 0$ can be found by defining the Lagrangian $\mathcal{L}(x, \lambda) = f(x) + \lambda g(x)$:

\begin{equation}
    \begin{split}
        \mathcal{L}(\mathbf{p}, \lambda) &= -\sum_{i=1}^{|\mathcal{A}|}\left(p_i \log p_i\right) + \lambda \left(\sum_{i=1}^{|\mathcal{A}|}p_i - 1\right) \\
        &= - \sum_{i=1}^{|\mathcal{A}|} \left(p_i \log p_i - \lambda p_i\right) - \lambda\\
        &= - \sum_{i=1}^{|\mathcal{A}|} \left(\log p_i - \lambda\right) p_i - \lambda.
    \end{split}
\end{equation}

\noindent Now, we can compute its gradient with the product rule:

\begin{equation}
    \begin{split}
        \nabla \mathcal{L}(\mathbf{p}, \lambda) &= \nabla \left[- \sum_{i=1}^{|\mathcal{A}|} \left(\log p_i - \lambda\right) p_i \right] - \nabla \lambda\\
        &= - \sum_{i=1}^{|\mathcal{A}|} \left(\left(\log p_i - \lambda\right) \nabla p_i + p_i \nabla \left(\log p_i - \lambda\right)\right) - \nabla \lambda\\
        &= - \sum_{i=1}^{|\mathcal{A}|} \left(\left(\log p_i - \lambda\right) \nabla p_i + p_i \nabla \log p_i - p_i \nabla \lambda\right) - \nabla \lambda,
    \end{split}
\end{equation}

\noindent where $\nabla \lambda$ is null. Given that $\nabla \log p_i = \frac{\nabla p_i}{p_i}$, we can replace $\nabla \log p_i$ in the second term, obtaining:

\begin{equation}
    \begin{split}
        \nabla \mathcal{L}(\mathbf{p}, \lambda) &= - \sum_{i=1}^{|\mathcal{A}|} \left(\left(\log p_i - \lambda\right) \nabla p_i + p_i \frac{\nabla p_i}{p_i}\right) \\
        &= - \sum_{i=1}^{|\mathcal{A}|} \left(\left(\log p_i - \lambda\right) \nabla p_i + \nabla p_i\right) \\
        &= - \sum_{i=1}^{|\mathcal{A}|} \left(\log p_i + 1 - \lambda\right) \nabla p_i.
    \end{split}
\end{equation}

Since the aim is to maximize entropy, we need to find where all partial derivatives are equal to 0, i.e., when the original constraint $g(x) = \sum_{i=1}^{|\mathcal{A}|} p_i - 1 = 0$ is satisfied, and when $\log p_i + 1 - \lambda = 0, \forall \, i \in \left[1, |\mathcal{A}|\right]$. The last set of equations shows that all $p_i$ must be equal; given the original constraint, this means that $p_i = \frac{1}{|\mathcal{A}|}, \forall \, i \in \left[1, |\mathcal{A}|\right]$, and that $\lambda = 1 + \log \frac{1}{|\mathcal{A}|} = 1 + \log 1 - \log |\mathcal{A}| = 1 - \log |\mathcal{A}|$. This stationary point represents the unique global maximum, since $H$ is strictly concave and the constraint is linear. When $\log p_i < \lambda - 1$, i.e., $p_i < \frac{1}{|\mathcal{A}|}$, the sign of the gradient is positive, and the probability is increased; vice versa, when $\log p_i > \lambda - 1$, the sign of the gradient becomes negative, and the probability is decreased.

Disequilibrium is defined as $D[\mathbf{p}] = \sum_{i=1}^{|\mathcal{A}|} \left(p_i - \frac{1}{|\mathcal{A}|}\right)^2$ and subject to the probability constraint $\sum_{i=1}^{|\mathcal{A}|} p_i = 1$. Again, we can make use of the method of Lagrange multipliers, obtaining:

\begin{equation}
    \begin{split}
        \mathcal{L}(\mathbf{p}, \lambda) &= \sum_{i=1}^{|\mathcal{A}|} \left(p_i - \frac{1}{|\mathcal{A}|}\right)^2 + \lambda \left(\sum_{i=1}^{|\mathcal{A}|} p_i - 1\right) \\
        &= \sum_{i=1}^{|\mathcal{A}|} \left(p_i^2 + \frac{1}{|\mathcal{A}|^2} - \frac{2}{|\mathcal{A}|}p_i \right) + \lambda \sum_{i=1}^{|\mathcal{A}|} p_i - \lambda \\
        &= \sum_{i=1}^{|\mathcal{A}|} \left(p_i^2 - \frac{2}{|\mathcal{A}|} p_i + \lambda p_i \right) + \frac{1}{|\mathcal{A}|} - \lambda \\
        &= \sum_{i=1}^{|\mathcal{A}|} \left(p_i^2 + \frac{\lambda |\mathcal{A}| - 2}{|\mathcal{A}|} p_i\right) + \frac{1}{|\mathcal{A}|} - \lambda.
    \end{split}
\end{equation}

\noindent Its gradient with respect to $\mathbf{p}$ can be defined as follows:

\begin{equation}
    \begin{split}
        \nabla \mathcal{L}(\mathbf{p}, \lambda) &= \sum_{i=1}^{|\mathcal{A}|} \left(2 p_i \nabla p_i + \frac{\lambda |\mathcal{A}| - 2}{|\mathcal{A}|} \nabla p_i \right) \\
        &= \sum_{i=1}^{|\mathcal{A}|} \left(2 p_i + \frac{\lambda |\mathcal{A}| - 2}{|\mathcal{A}|}\right) \nabla p_i.
    \end{split}
\end{equation}

\noindent To find the optimum, we need to set all partial derivatives equal to $0$, including with respect to $\lambda$, i.e., that the original constraint $g(x) = \sum_{i=1}^{|\mathcal{A}|} p_i - 1 = 0$ is satisfied, plus that $2 p_i + \frac{\lambda |\mathcal{A}| - 2}{|\mathcal{A}|} = 0, \forall \, i \in \left[1, |\mathcal{A}|\right]$. This set of equations is satisfied for $p_i = \frac{1}{|\mathcal{A}|} - \frac{\lambda}{2}$. But since we also have the original constraint, $\sum_{i=1}^{|\mathcal{A}|} \left(\frac{1}{|\mathcal{A}|} - \frac{\lambda}{2}\right) - 1 = 1 - \frac{\lambda |\mathcal{A}|}{2} -1 = - \frac{\lambda |\mathcal{A}|}{2}$ must be equal to zero; thus, $\lambda = 0$ and $p_i = \frac{1}{|\mathcal{A}|}, \forall \, i \in \left[1, |\mathcal{A}|\right]$. As expected, the minimum lies where the distribution is equiprobable. When $p_i < \frac{1}{|\mathcal{A}|}$, the gradient is negative, causing the probability $p_i$ to decrease further. Conversely, when $p_i > \frac{1}{|\mathcal{A}|}$, the gradient becomes positive, leading to an increase in the probability $p_i$.

Finally, complexity is defined as $C[\mathbf{p}] = H[\mathbf{p}] \cdot D[\mathbf{p}]$. According to the product rule, its gradient is as follows:

\begin{equation}
    \begin{split}
        \nabla &C[\mathbf{p}] =  H[\mathbf{p}] \nabla D[\mathbf{p}] + D[\mathbf{p}] \nabla H[\mathbf{p}] \\
        &= \sum_{i=1}^{|\mathcal{A}|} \left[ 2 H[\mathbf{p}] \left( p_i - \frac{1}{|\mathcal{A}|} \right) - D[\mathbf{p}] \left( \log p_i + 1 \right) \right] \nabla p_i \\
        &= \sum_{i=1}^{|\mathcal{A}|} \left[ 2 H[\mathbf{p}] p_i - \frac{2}{|\mathcal{A}|} H[\mathbf{p}] - D[\mathbf{p}] \log p_i - D[\mathbf{p}] \right] \nabla p_i.
    \end{split}
\end{equation}

However, as for entropy and disequilibrium, complexity is subject to the probability constraint $\sum_{i=1}^{|\mathcal{A}|} p_i = 1$ as well. To find the global optima of complexity, we need to find where all the partial derivatives of the Lagrangian $C[\mathbf{p}] + \lambda \left(\sum_{i=1}^{|\mathcal{A}|} p_i - 1\right)$ are equal to $0$, i.e., to solve the equation system composed of the probability constraint plus $2H[\mathbf{p}] p_i - \frac{2}{|\mathcal{A}|} H[\mathbf{p}] - D[\mathbf{p}] \log p_i - D[\mathbf{p}] + \lambda = 0, \forall \, i \in \left[1, |\mathcal{A}|\right]$. This system has multiple solutions, i.e., where $p_i = \frac{1}{|\mathcal{A}|}, \forall \, i \in \left[1, |\mathcal{A}|\right]$ (which represents a global minimum and complexity is equal to 0; also the deterministic vertices are global minima with complexity equal to 0, but they are outside the Lagrangian analysis since $\log 0$ is undefined), plus $|\mathcal{A}|$ solutions, which correspond to the complexity maxima, which are the recombination of the same set of probabilities where one is equal to a (large) value $p$ and all the others are equal to  $\frac{1 - p}{|\mathcal{A}|-1}$ \citep{anteneodo1996some,calbet2001tendency}.

\section{Implementation Details} \label{sec:implementation_details}

The experiments were carried out on a system equipped with a 32-core AMD EPYC 7413 processor and an NVIDIA L40 GPU, running Python version 3.10.18. We used the \texttt{torch} library and leveraged the \texttt{stable-baselines3} implementation of PPO, which was modified into CR-PPO according to Algorithm \ref{alg:crppo_python}. The complete implementation, experimental scripts and plot generation are available at \url{https://github.com/Lucaserf/CR-PPO}.

\begin{algbox}[label={alg:crppo_python}, colback=white, colframe=black!80, left=2pt, boxsep=2pt, boxrule=1.0pt]{Python implementation of CR-PPO complexity loss}
$\cdot \cdot \cdot$ \\
$\mathtt{dist = self.policy.get\_distribution(rollout\_data.observations)}$ \\
$\mathtt{probs = dist.distribution.probs}$ \\
$\mathtt{entropy = dist.distribution.entropy()}$ \\
$\mathtt{disequilibrium = torch.sum((probs - 1.0 / probs.size(-1))^{**}2, dim=-1)}$ \\
$\mathtt{complexity = entropy * disequilibrium}$ \\
$\mathtt{complexity\_loss = - torch.mean(complexity)}$ \\
$\cdot \cdot \cdot$
\end{algbox}

The adopted neural networks followed the default architectures provided by \texttt{stable-baselines3}. For 1D observation space environments, i.e., CartPole and CARTerpillar environments, we used an MLP with 2 fully connected layers with 64 units each. For image observation spaces, i.e., CarRacing, Crafter, and Atari environments, we used a Nature CNN architecture for feature extraction, with a sequence of 2D convolutional layers with layer sizes $[32, 64, 64]$, kernel sizes $[(8,8),(4,4),(3,3)]$, and strides $[(4,4),(2,2),(1,1)]$, with ReLU activation function. The output is then flattened and passed through a dense layer (of dimension 256 for CarRacing and 512 for Crafter and Atari games) with ReLU activation. 
Finally, consistent with the original CoinRun experiments \citep{cobbe2020leveraging}, we used the IMPALA CNN architecture \citep{espeholt2018impala} with three layers, each consisting of a 2D convolutional layer, a 2D max-pooling operation with kernel size 3 and stride 2, followed by two residual blocks. Each residual block contains two 2D convolutional layers preceded by ReLU activation functions, and their output is summed with their input. The convolutional layers inside the three IMPALA layers have the size ($[16, 32, 32]$, respectively), and the same stride of 3. Finally, the output is flattened and passed through a dense layer of dimension 256 with ReLU activation.

The full list of training hyperparameters is reported in Table \ref{tab:hyperparameters}.

\begin{table}[ht]
    \caption{Training hyperparameters for all the experiments.}
    \label{tab:hyperparameters}
    \centering
    \resizebox{\textwidth}{!}{%
    \begin{tabular}{lccccc}
        \hline \noalign{\vskip 1mm}
        \textbf{Parameter} & \textbf{CartPole} & \textbf{CarRacing} & \textbf{CoinRun} & \textbf{Crafter \& Atari games} & \textbf{CARTerpillar} \\
        \hline \noalign{\vskip 1mm}
        Number of parallel envs & 8 & 8 & 256 & 8 & 8 \\
        Steps between updates & 32 & 512 & 256 & 128 & 32 \\
        Policy epochs & 20 & 20 & 3 & 4 & 20 \\
        Batch size & 256 & 128 & 2048 & 256 & 256 \\
        GAE lambda & 0.8 & 0.95 & 0.95 & 0.95 & 0.8 \\
        Gamma & 0.98 & 0.99 & 0.99 & 0.99 & 0.98 \\
        Max gradient norm & 0.5 & 0.5 & 0.5 & 0.5 & 0.5 \\
        Learning rate & 1e-3 & 1e-4 & 5e-4 & 2.5e-4 & 1e-3 \\
        Clip range $\epsilon$ & 0.2 & 0.2 & 0.2 & 0.1 & 0.2 \\
        Value-function coeff. $c_{vf}$ & 0.5 & 0.5 & 0.5 & 0.5 & 0.5 \\
        \noalign{\vskip 1mm} 
        \hline \noalign{\vskip 1mm}
        Max episode steps & 500 & - & - & - & 500 \\
        Gravity constant $G$ & 9.8 & - & - & - & 9.81 \\
        Frame skip & - & 2 & - & - & - \\
        Frame stack & - & 2 & - & 4 & - \\
        SDE sample frequency & - & 4 & - & - & - \\
        Number of parallel workers & - & - & 2 & - & - \\
        Number of training levels & - & - & 500 & - & - \\
        Number of testing levels & - & - & $\infty$ & - & - \\
        Number of minibatches & - & - & 8 & - & - \\
        Distribution mode & - & - & Hard & - & - \\
        Value function clipping & - & - & True & - & - \\
        Spring constant $k$ & - & - & - & - & 1 \\
        Damper constant $b$ & - & - & - & - & 1 \\
        \hline
    \end{tabular}
    }
\end{table}

\section{CARTerpillar Additional Results} \label{sec:additional_results}

\begin{figure}[ht]
    \centering
    \includegraphics[width=1\linewidth]{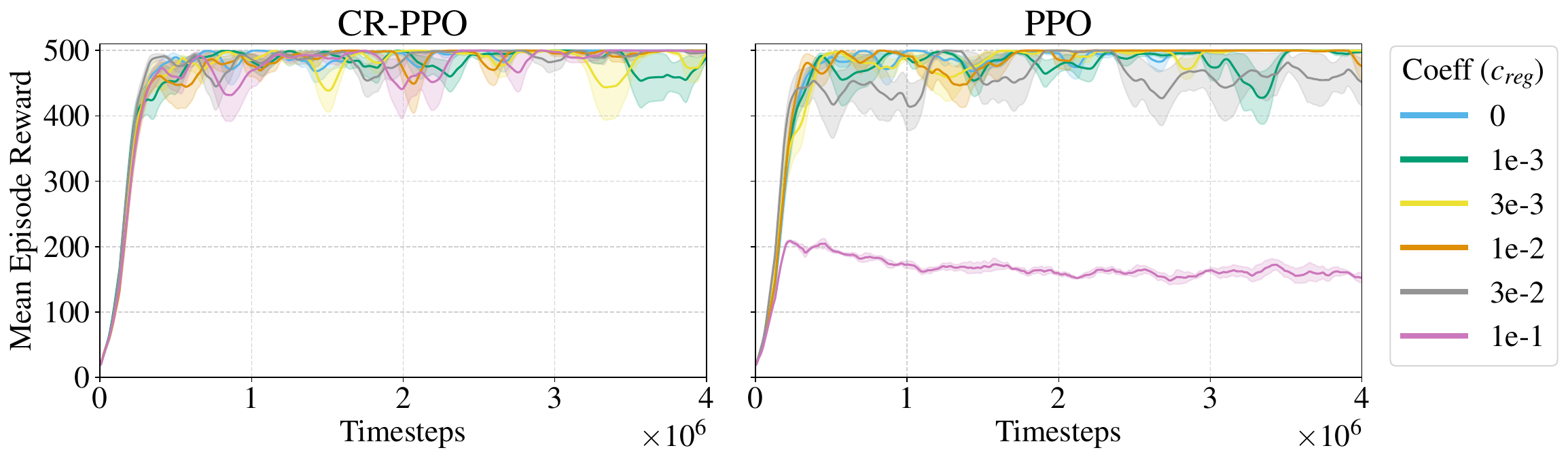}
    \caption{Mean episode reward for CARTerpillar with $K=6$ carts for CR-PPO (left) and PPOwEnt (right) with different $c_{reg}$ values. Each plot reports the mean and standard error over 5 seeds.}
    \label{fig:cart6}
\end{figure}

\begin{figure}[ht]
    \centering
    \includegraphics[width=1\linewidth]{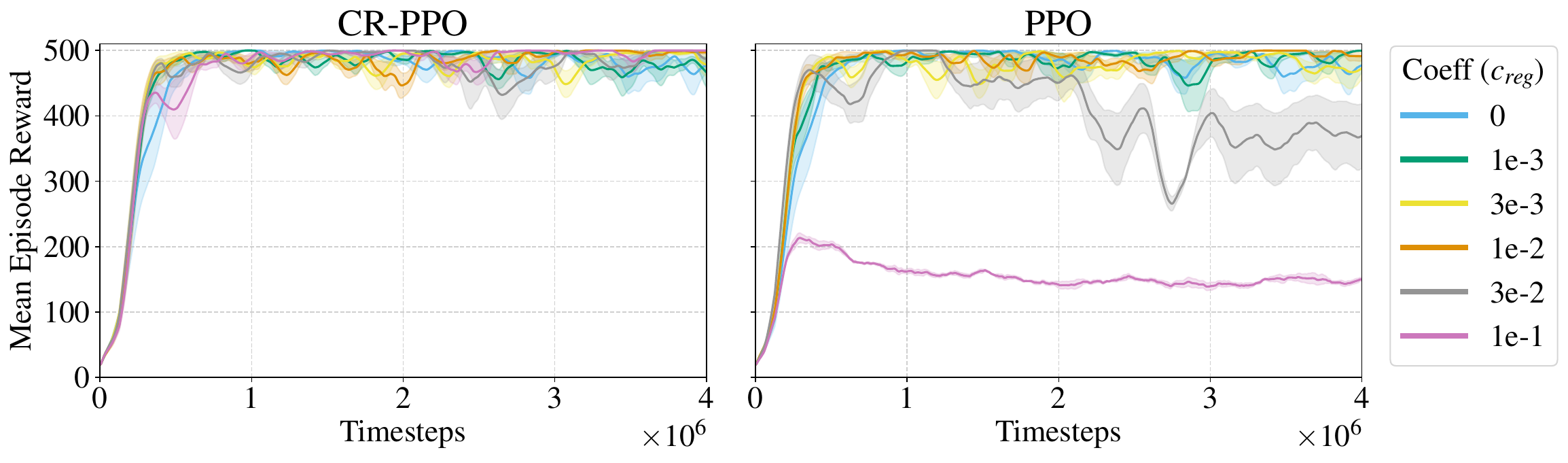}
    \caption{Mean episode reward for CARTerpillar with $K=7$ carts for CR-PPO (left) and PPOwEnt (right) with different $c_{reg}$ values. Each plot reports the mean and standard error over 5 seeds.}
    \label{fig:cart7}
\end{figure}

\begin{figure}[ht]
    \centering
    \includegraphics[width=1\linewidth]{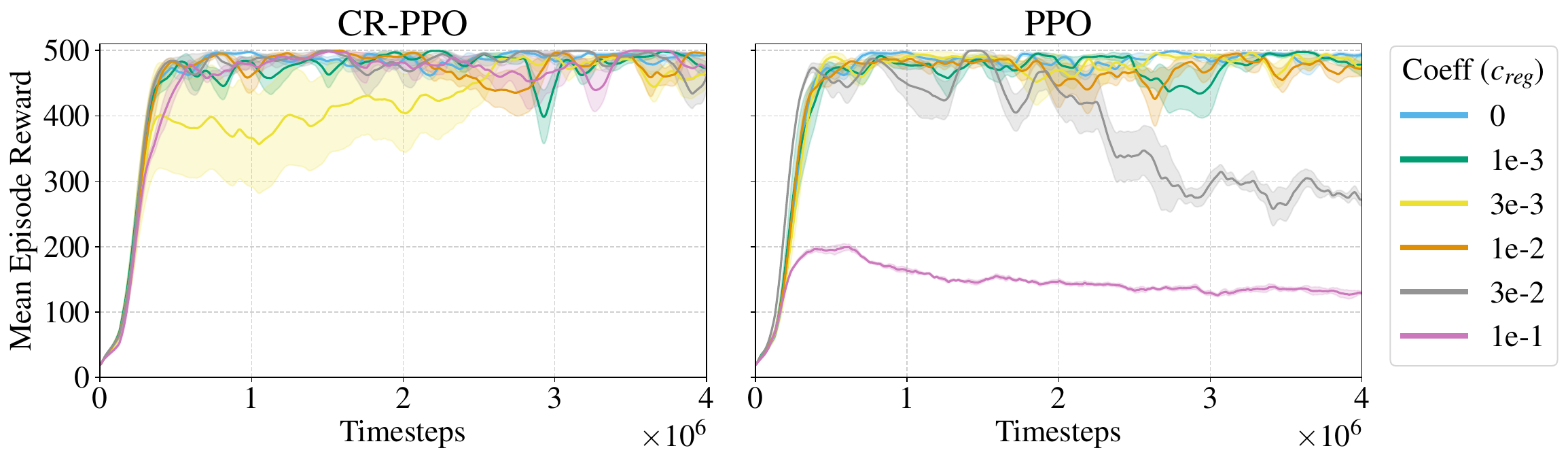}
    \caption{Mean episode reward for CARTerpillar with $K=8$ carts for CR-PPO (left) and PPOwEnt (right) with different $c_{reg}$ values. Each plot reports the mean and standard error over 5 seeds.}
    \label{fig:cart8}
\end{figure}

\begin{figure}[ht]
    \centering
    \includegraphics[width=1\linewidth]{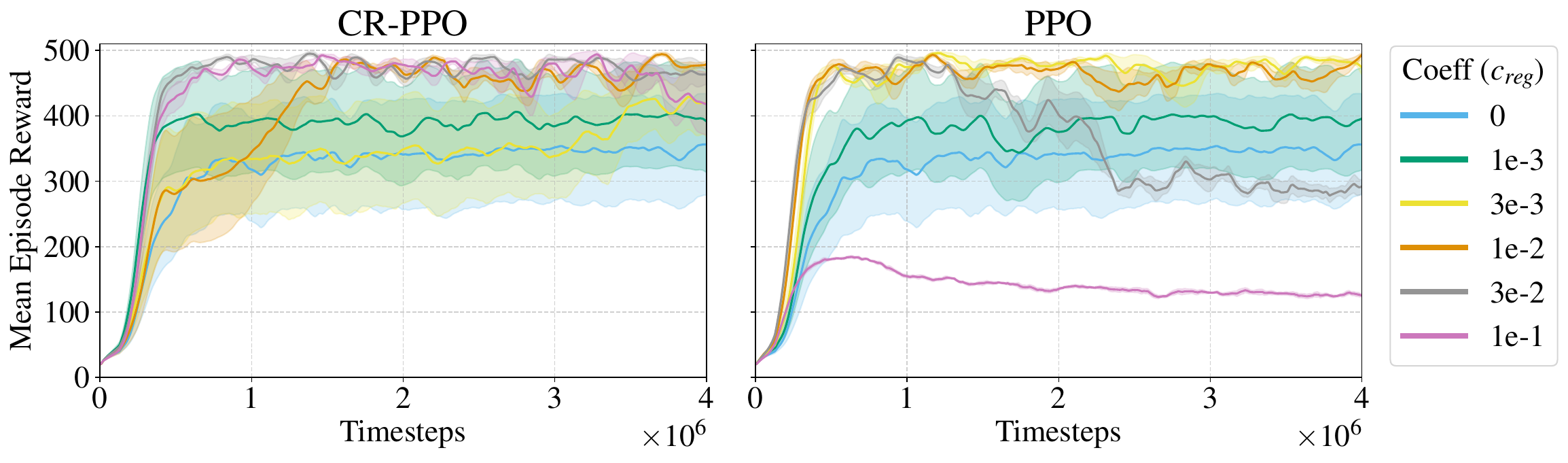}
    \caption{Mean episode reward for CARTerpillar with $K=9$ carts for CR-PPO (left) and PPOwEnt (right) with different $c_{reg}$ values. Each plot reports the mean and standard error over 5 seeds.}
    \label{fig:cart9}
\end{figure}

\begin{figure}[ht]
    \centering
    \includegraphics[width=1\linewidth]{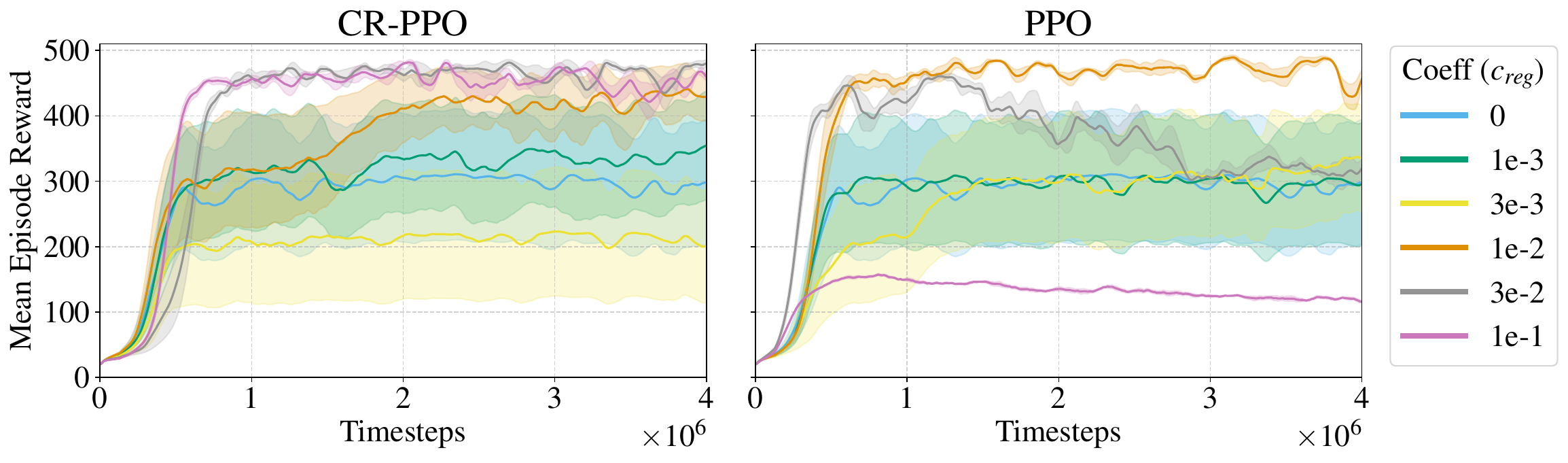}
    \caption{Mean episode reward for CARTerpillar with $K=10$ carts for CR-PPO (left) and PPOwEnt (right) with different $c_{reg}$ values. Each plot reports the mean and standard error over 5 seeds.}
    \label{fig:cart10}
\end{figure}

\begin{figure}[ht]
    \centering
    \includegraphics[width=1\linewidth]{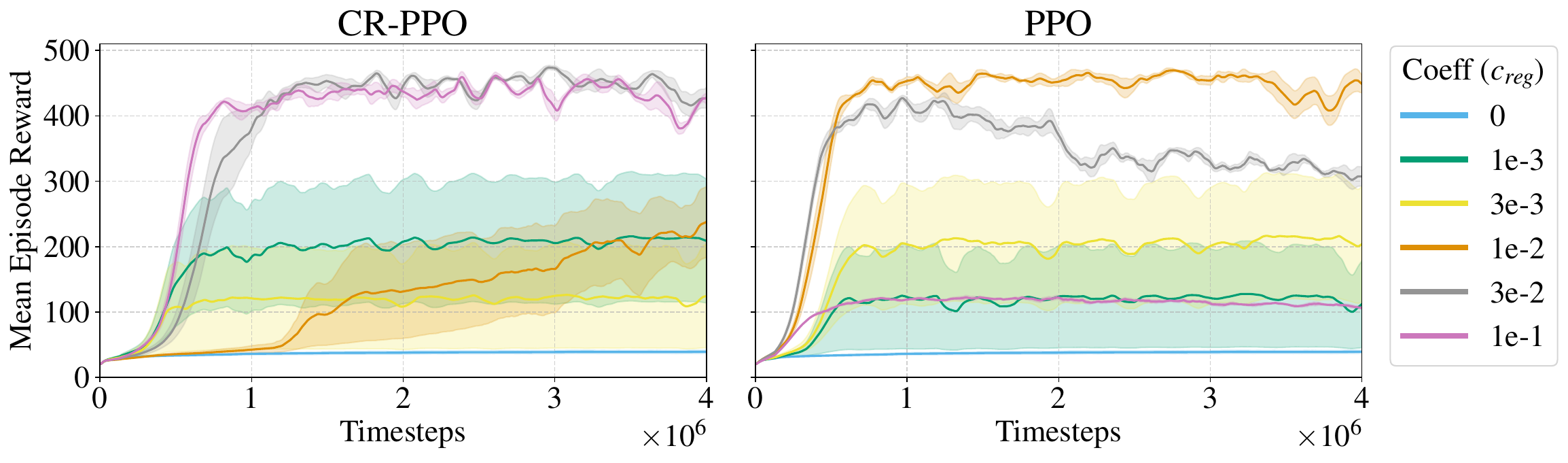}
    \caption{Mean episode reward for CARTerpillar with $K=11$ carts for CR-PPO (left) and PPOwEnt (right) with different $c_{reg}$ values. Each plot reports the mean and standard error over 5 seeds.}
    \label{fig:cart11}
\end{figure}

\begin{figure}[ht]
    \centering
    \includegraphics[width=1\linewidth]{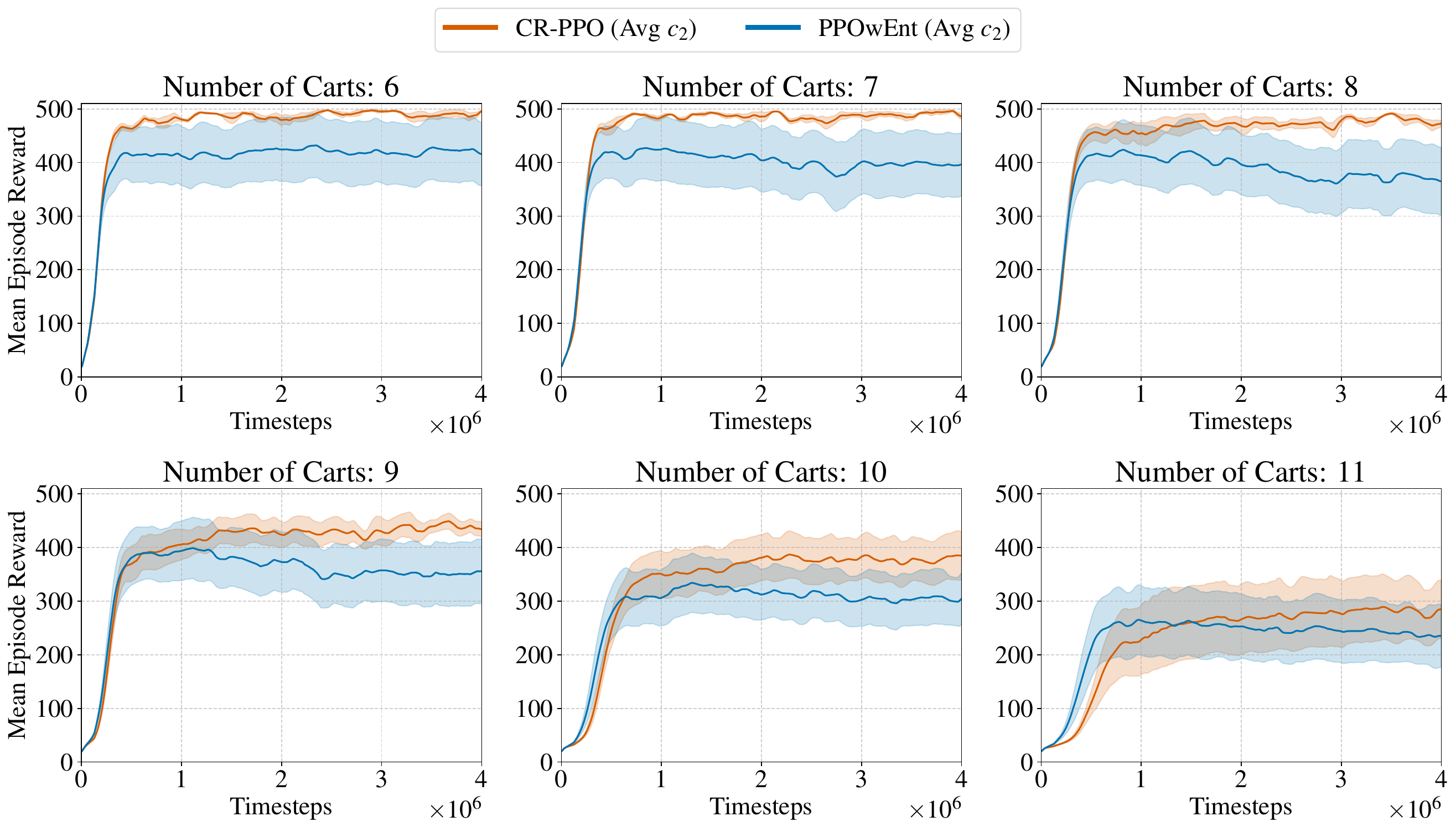}
    \caption{Aggregated results for CR-PPO and PPOwEnt over different $c_{reg}$ values for the CARTerpillar environment with varying number of carts. Each plot reports the mean and standard error over 5 seeds.}
    \label{fig:carterpillar}
\end{figure}

In the following, we report the detailed results of CR-PPO and PPOwEnt for different CARTerpillar configurations. In particular, we evaluate their performances for five $c_{reg}$ values: $[1\text{e-}1, 3\text{e-}2, 1\text{e-}2, 3\text{e-}3, 1\text{e-}3]$. For 6, 7, and 8 carts (Figures \ref{fig:cart6}, \ref{fig:cart7}, and \ref{fig:cart8}, respectively), all $c_{reg}$ values reach the optimum in the case of CR-PPO, while some of them struggle in the case of PPOwEnt. The same holds for 9 carts (Figure \ref{fig:cart9}), where a too-high entropy coefficient can make the agent unlearn optimal strategies. Finally, only a well-tuned coefficient can make PPOwEnt converge for 10 and 11 carts (Figures \ref{fig:cart10} and \ref{fig:cart11}), while CR-PPO can converge with multiple $c_{reg}$ values. This greater robustness to variations in $c_{reg}$ is demonstrated in Figure \ref{fig:carterpillar}, which presents the mean and standard error of the aggregated results for all evaluated cart numbers.

\end{document}

%% file: math_commands.tex
\usepackage{amsmath,amsfonts,bm}

\def\eqref#1{equation~\ref{#1}}

\def\1{\bm{1}}

\DeclareMathAlphabet{\mathsfit}{\encodingdefault}{\sfdefault}{m}{sl}
\SetMathAlphabet{\mathsfit}{bold}{\encodingdefault}{\sfdefault}{bx}{n}

\newcommand{\E}{\mathbb{E}}



%% file: biblio.bib
@article{lopezruiz1995statistical,
  title   = {A statistical measure of complexity},
  journal = {Physics Letters A},
  volume  = {209},
  number  = {5},
  pages   = {321-326},
  year    = {1995},
  author  = {Ricardo López-Ruiz and Hector L. Mancini and Xavier Calbet}
}

@inproceedings{cobbe2020leveraging,
  author    = {Cobbe, Karl and Hesse, Christopher and Hilton, Jacob and Schulman, John},
  title     = {Leveraging procedural generation to benchmark reinforcement learning},
  year      = {2020},
  booktitle = {{Proc. of the 37th International Conference on Machine Learning (ICML'20)}}
}

@misc{raffin2021rl,
  author       = {Raffin, Antonin},
  title        = {{RL Baselines3 Zoo}},
  year         = {2020},
  publisher    = {GitHub},
  journal      = {GitHub repository},
  howpublished = {\url{https://github.com/DLR-RM/rl-baselines3-zoo}},
  note         = {Accessed: 2026-11-22}
}

@inproceedings{towers2025gymnasium,
  title     = {Gymnasium: A Standard Interface for Reinforcement Learning Environments},
  author    = {Mark Towers and Ariel Kwiatkowski and John U. Balis and Gianluca De Cola and Tristan Deleu and Manuel Goul{\~a}o and Kallinteris Andreas and Markus Krimmel and Arjun KG and Rodrigo De Lazcano Perez-Vicente and J K Terry and Andrea Pierr{\'e} and Sander V Schulhoff and Jun Jet Tai and Hannah Tan and Omar G. Younis},
  booktitle = {{Proc. of the NeurIPS'25 Datasets and Benchmarks Track}},
  year      = {2025}
}

@inproceedings{sutton1999policy,
  author    = {Sutton, Richard S. and McAllester, David and Singh, Satinder and Mansour, Yishay},
  booktitle = {{Advances in Neural Information Processing Systems (NeurIPS'99)}},
  title     = {Policy Gradient Methods for Reinforcement Learning with Function Approximation},
  year      = {1999}
}

@misc{schulman2017proximal,
  title  = {Proximal Policy Optimization Algorithms},
  author = {Schulman, John and Wolski, Filip and Dhariwal, Prafulla and Radford, Alec and Klimov, Oleg},
  year   = {2017},
  note   = {arXiv:1707.06347 [cs.LG]}
}

@inproceedings{schulman2016highdimensional,
  author    = {Schulman, John and Moritz, Philipp and Levine, Sergey and Jordan, Michael I. and Abbeel, Pieter},
  title     = {High-Dimensional Continuous Control Using Generalized Advantage Estimation},
  booktitle = {{Proc. of the 4th International Conference on Learning Representations (ICLR'16)}},
  year      = {2016}
}

@inproceedings{zhang2025intelligence,
  title     = {Intelligence at the Edge of Chaos},
  author    = {Shiyang Zhang and Aakash Patel and Syed A Rizvi and Nianchen Liu and Sizhuang He and Amin Karbasi and Emanuele Zappala and David van Dijk},
  booktitle = {{Proc. of the 13th International Conference on Learning Representations (ICLR'25)}},
  year      = {2025}
}

@inproceedings{young2024enhancing,
  title     = {Enhancing Robustness in Deep Reinforcement Learning: A Lyapunov Exponent Approach},
  author    = {Rory Young and Nicolas Pugeault},
  booktitle = {{Proc. of the 38th Annual Conference on Neural Information Processing Systems (NeurIPS'24)}},
  year      = {2024}
}

@misc{zhang2021edge,
  title  = {Edge of chaos as a guiding principle for modern neural network training},
  author = {Lin Zhang and Ling Feng and Kan Chen and Choy Heng Lai},
  year   = {2021},
  note   = {arXiv:2107.09437 [cs.LG]}
}

@article{lempel1976complexity,
  author  = {Lempel, Abraham and Ziv, Jacob},
  journal = {IEEE Transactions on Information Theory},
  title   = {On the Complexity of Finite Sequences},
  year    = {1976},
  volume  = {22},
  number  = {1},
  pages   = {75-81}
}

@article{grassberger1986toward,
  author  = {Grassberger, Peter},
  title   = {Toward a quantitative theory of self-generated complexity},
  journal = {International Journal of Theoretical Physics},
  year    = {1986},
  volume  = {25},
  number  = {9},
  pages   = {907-938}
}

@inproceedings{zhang2025when,
  title     = {When Maximum Entropy Misleads Policy Optimization},
  author    = {Ruipeng Zhang and Ya-Chien Chang and Sicun Gao},
  booktitle = {{Proc. of the 42nd International Conference on Machine Learning (ICML'25)}},
  year      = {2025}
}

@inproceedings{ahmed2019understanding,
  title     = {Understanding the Impact of Entropy on Policy Optimization},
  author    = {Ahmed, Zafarali and Le Roux, Nicolas and Norouzi, Mohammad and Schuurmans, Dale},
  booktitle = {{Proc. of the 36th International Conference on Machine Learning (ICML'19)}},
  year      = {2019}
}

@inproceedings{haarnoja2018soft,
  title     = {Soft Actor-Critic: Off-Policy Maximum Entropy Deep Reinforcement Learning with a Stochastic Actor},
  author    = {Haarnoja, Tuomas and Zhou, Aurick and Abbeel, Pieter and Levine, Sergey},
  booktitle = {{Proc. of the 35th International Conference on Machine Learning (ICML'18)}},
  year      = {2018}
}

@inproceedings{andrychowicz2021what,
  title     = {What Matters for On-Policy Deep Actor-Critic Methods? A Large-Scale Study},
  author    = {Marcin Andrychowicz and Anton Raichuk and Piotr Sta{\'n}czyk and Manu Orsini and Sertan Girgin and Rapha{\"e}l Marinier and Leonard Hussenot and Matthieu Geist and Olivier Pietquin and Marcin Michalski and Sylvain Gelly and Olivier Bachem},
  booktitle = {{Proc. of the 9th International Conference on Learning Representations (ICLR'21)}},
  year      = {2021}
}

@article{mnih2015humanlevel,
  title   = {Human-Level Control Through Deep Reinforcement Learning},
  author  = {Mnih, Volodymyr and Kavukcuoglu, Koray and Silver, David and Rusu, Andrei A. and Veness, Joel and Bellemare, Marc G. and Graves, Alex and Riedmiller, Martin and Fidjeland, Andreas K. and Ostrovski, Georg and Petersen, Stig and Beattie, Charles and Sadik, Amir and Antonoglou, Ioannis and King, Helen and Kumaran, Dharshan and Wierstra, Daan and Legg, Shane and Hassabis, Demis},
  journal = {Nature},
  volume  = {518},
  pages   = {529-533},
  year    = {2015}
}

@article{silver2017mastering,
  author  = {Silver, David and Schrittwieser, Julian and Simonyan, Karen and Antonoglou, Ioannis and Huang, Aja and Guez, Arthur and Hubert, Thomas and Baker, Lucas and Lai, Matthew and Bolton, Adrian and Chen, Yutian and Lillicrap, Timothy and Hui, Fan and Sifre, Laurent and van den Driessche, George and Graepel, Thore and Hassabis, Demis},
  title   = {Mastering the game of Go without human knowledge},
  journal = {Nature},
  year    = {2017},
  volume  = {550},
  number  = {7676},
  pages   = {354-359}
}

@inproceedings{angermueller2020modelbased,
  title     = {Model-based reinforcement learning for biological sequence design},
  author    = {Christof Angermueller and David Dohan and David Belanger and Ramya Deshpande and Kevin Murphy and Lucy Colwell},
  booktitle = {{Proc. of the 8th International Conference on Learning Representations (ICLR'20)}},
  year      = {2020}
}

@inproceedings{ouyang2022training,
  author    = {Ouyang, Long and Wu, Jeffrey and Jiang, Xu and Almeida, Diogo and Wainwright, Carroll and Mishkin, Pamela and Zhang, Chong and Agarwal, Sandhini and Slama, Katarina and Ray, Alex and Schulman, John and Hilton, Jacob and Kelton, Fraser and Miller, Luke and Simens, Maddie and Askell, Amanda and Welinder, Peter and Christiano, Paul F and Leike, Jan and Lowe, Ryan},
  booktitle = {{Advances in Neural Information Processing Systems (NeurIPS'22)}},
  title     = {Training language models to follow instructions with human feedback},
  year      = {2022}
}

@book{sutton2018reinforcement,
  title     = {Reinforcement Learning: An Introduction},
  author    = {Sutton, Richard S. and Barto, Andrew G.},
  year      = {2018},
  publisher = {MIT Press}
}

@article{tang2025deep,
  author  = {Tang, Chen and Abbatematteo, Ben and Hu, Jiaheng and Chandra, Rohan and Martín-Martín, Roberto and Stone, Peter},
  title   = {Deep Reinforcement Learning for Robotics: A Survey of Real-World Successes},
  journal = {Annual Review of Control, Robotics, and Autonomous Systems},
  year    = {2025},
  volume  = {8},
  pages   = {153-188}
}

@misc{schulman2017equivalence,
  title  = {Equivalence Between Policy Gradients and Soft Q-Learning},
  author = {John Schulman and Xi Chen and Pieter Abbeel},
  year   = {2017},
  note   = {arXiv:1704.06440 [cs.LG]}
}

@article{williams1991function,
  author  = {Ronald J. Williams and Jing Peng},
  title   = {Function Optimization using Connectionist Reinforcement Learning Algorithms},
  journal = {Connection Science},
  volume  = {3},
  number  = {3},
  pages   = {241-268},
  year    = {1991}
}

@inproceedings{odonoghue2017combining,
  title     = {Combining policy gradient and Q-learning},
  author    = {Brendan O'Donoghue and Remi Munos and Koray Kavukcuoglu and Volodymyr Mnih},
  booktitle = {{Proc. of the 5th International Conference on Learning Representations (ICLR'17)}},
  year      = {2017}
}

@inproceedings{mnih2016asynchronous,
  title     = {Asynchronous Methods for Deep Reinforcement Learning},
  author    = {Mnih, Volodymyr and Badia, Adria Puigdomenech and Mirza, Mehdi and Graves, Alex and Lillicrap, Timothy and Harley, Tim and Silver, David and Kavukcuoglu, Koray},
  booktitle = {{Proc. of the 33rd International Conference on Machine Learning (ICML'16)}},
  year      = {2016}
}

@inproceedings{zhao2019maximum,
  title     = {Maximum Entropy-Regularized Multi-Goal Reinforcement Learning},
  author    = {Zhao, Rui and Sun, Xudong and Tresp, Volker},
  booktitle = {{Proc. of the 36th International Conference on Machine Learning (ICML'19)}},
  year      = {2019}
}

@inproceedings{han2021maxmin,
  author    = {Han, Seungyul and Sung, Youngchul},
  booktitle = {{Advances in Neural Information Processing Systems (NeurIPS'21)}},
  title     = {A Max-Min Entropy Framework for Reinforcement Learning},
  year      = {2021}
}

@misc{neu2017unified,
  title  = {{A unified view of entropy-regularized Markov decision processes}},
  author = {Gergely Neu and Anders Jonsson and Vicenç Gómez},
  year   = {2017},
  note   = {arXiv:1705.07798 [cs.LG]}
}

@inproceedings{haarnoja2017reinforcement,
  title     = {Reinforcement Learning with Deep Energy-Based Policies},
  author    = {Tuomas Haarnoja and Haoran Tang and Pieter Abbeel and Sergey Levine},
  booktitle = {Proc. of the 34th International Conference on Machine Learning (ICML'17)},
  year      = {2017}
}

@inproceedings{ziebart2008maximum,
  author    = {Ziebart, Brian D. and Maas, Andrew and Bagnell, J.Andrew and Dey, Anind K.},
  title     = {Maximum Entropy Inverse Reinforcement Learning},
  booktitle = {{Proc. of the 23rd AAAI Conference on Artificial Intelligence (AAAI'08)}},
  year      = {2008}
}

@inproceedings{toussaint2009robot,
  author    = {Toussaint, Marc},
  title     = {Robot trajectory optimization using approximate inference},
  year      = {2009},
  booktitle = {{Proc. of the 26th Annual International Conference on Machine Learning (ICML'09)}}
}

@inproceedings{rawlik2013stochastic,
  author    = {Rawlik, Konrad and Toussaint, Marc and Vijayakumar, Sethu},
  title     = {On stochastic optimal control and reinforcement learning by approximate inference},
  year      = {2013},
  booktitle = {{Proc. of the 23rd International Joint Conference on Artificial Intelligence (IJCAI'13)}}
}

@inproceedings{han2021diversity,
  title     = {Diversity Actor-Critic: Sample-Aware Entropy Regularization for Sample-Efficient Exploration},
  author    = {Han, Seungyul and Sung, Youngchul},
  booktitle = {{Proc. of the 38th International Conference on Machine Learning (ICML'21)}},
  year      = {2021}
}

@inproceedings{liu2021regularization,
  title     = {Regularization Matters in Policy Optimization - An Empirical Study on Continuous Control},
  author    = {Zhuang Liu and Xuanlin Li and Bingyi Kang and Trevor Darrell},
  booktitle = {{Proc. of the 9th International Conference on Learning Representations (ICLR'21)}},
  year      = {2021}
}

@inproceedings{hazan2019provably,
  title     = {Provably Efficient Maximum Entropy Exploration},
  author    = {Hazan, Elad and Kakade, Sham and Singh, Karan and Van Soest, Abby},
  booktitle = {{Proc. of the 36th International Conference on Machine Learning (ICML'19)}},
  year      = {2019}
}

@article{lyapunov1992general,
  author    = {Lyapunov, Aleksandr M.},
  title     = {The general problem of the stability of motion},
  journal   = {International Journal of Control},
  volume    = {55},
  number    = {3},
  pages     = {531-534},
  year      = {1992},
  publisher = {Taylor \& Francis}
}

@article{langton1990computation,
  title   = {Computation at the edge of chaos: Phase transitions and emergent computation},
  journal = {Physica D: Nonlinear Phenomena},
  volume  = {42},
  number  = {1},
  pages   = {12-37},
  year    = {1990},
  author  = {Chris G. Langton}
}

@misc{havrilla2024surveying,
  title  = {Surveying the Effects of Quality, Diversity, and Complexity in Synthetic Data From Large Language Models},
  author = {Alex Havrilla and Andrew Dai and Laura O'Mahony and Koen Oostermeijer and Vera Zisler and Alon Albalak and Fabrizio Milo and Sharath Chandra Raparthy and Kanishk Gandhi and Baber Abbasi and Duy Phung and Maia Iyer and Dakota Mahan and Chase Blagden and Srishti Gureja and Mohammed Hamdy and Wen-Ding Li and Giovanni Paolini and Pawan Sasanka Ammanamanchi and Elliot Meyerson},
  year   = {2024},
  note   = {arXiv:2412.02980 [cs.LG]}
}

@book{mitchell2009complexity,
  author    = {Mitchell, Melanie},
  title     = {{Complexity: A Guided Tour}},
  publisher = {Oxford University Press},
  year      = {2009}
}

@inproceedings{espeholt2018impala,
  title     = {{IMPALA}: Scalable Distributed Deep-{RL} with Importance Weighted Actor-Learner Architectures},
  author    = {Espeholt, Lasse and Soyer, Hubert and Munos, Remi and Simonyan, Karen and Mnih, Vlad and Ward, Tom and Doron, Yotam and Firoiu, Vlad and Harley, Tim and Dunning, Iain and Legg, Shane and Kavukcuoglu, Koray},
  booktitle = {{Proc. of the 35th International Conference on Machine Learning (ICML'18)}},
  year      = {2018}
}

@article{ladosz2022exploration,
  title   = {Exploration in deep reinforcement learning: A survey},
  journal = {Information Fusion},
  volume  = {85},
  pages   = {1-22},
  year    = {2022},
  author  = {Pawel Ladosz and Lilian Weng and Minwoo Kim and Hyondong Oh}
}

@inproceedings{hafner2022benchmarking,
  title     = {Benchmarking the Spectrum of Agent Capabilities},
  author    = {Danijar Hafner},
  booktitle = {{Proc. of the 10th International Conference on Learning Representations (ICLR'22)}},
  year      = {2022}
}

@article{calbet2001tendency,
    title = {Tendency towards maximum complexity in a nonequilibrium isolated system},
    author = {Calbet, Xavier and L{\'o}pez-Ruiz, Ricardo},
    journal = {Physical Review E},
    volume = {63},
    issue = {6},
    pages = {066116},
    year = {2001},
}

@article{anteneodo1996some,
    title = {Some features of the López-Ruiz-Mancini-Calbet (LMC) statistical measure of complexity},
    journal = {Physics Letters A},
    volume = {223},
    number = {5},
    pages = {348-354},
    year = {1996},
    author = {Anteneodo, Celia  and Plastino, \'{A}ngel Ricardo}
}
